\documentclass{article}

\PassOptionsToPackage{numbers, compress}{natbib}
\usepackage[preprint]{neurips_2022}
\usepackage{comment}

\usepackage{multirow}
\usepackage{amsmath}
\usepackage{subfig}
\usepackage[utf8]{inputenc} % allow utf-8 input
\usepackage[T1]{fontenc}    % use 8-bit T1 fonts
\usepackage{hyperref}       % hyperlinks
\usepackage{url}            % simple URL typesetting
\usepackage{booktabs}       % professional-quality tables
\usepackage{amsfonts}       % blackboard math symbols
\usepackage{nicefrac}       % compact symbols for 1/2, etc.
\usepackage{microtype}      % microtypography
\usepackage{xcolor}         % colors
\usepackage[pdftex]{graphicx}

\usepackage{gensymb}
\usepackage{booktabs}       % professional-quality tables
\usepackage{tabularx}
\usepackage{multirow}
\usepackage{multicol}
\newcommand{\PreserveBackslash}[1]{\let\temp=\\#1\let\\=\temp}
\newcolumntype{C}[1]{>{\PreserveBackslash\centering}p{#1}}
\newcolumntype{R}[1]{>{\PreserveBackslash\raggedleft}p{#1}}
\newcolumntype{L}[1]{>{\PreserveBackslash\raggedright}p{#1}}
\newcolumntype{?}{!{\vrule width 0.6pt}}
\title{Prototypical VoteNet for Few-Shot 3D Point Cloud Object Detection}

\author{
  Shizhen Zhao, Xiaojuan Qi\thanks{Corresponding author}\\
  The University of Hong Kong \\
  \texttt{\{zhaosz,xjqi\}@eee.hku.hk} \\
}

\begin{document}

\maketitle

\begin{abstract}

Most existing 3D point cloud object detection approaches heavily rely on large amounts of labeled training data.
However, the labeling process is costly and time-consuming.
This paper considers few-shot 3D point cloud object detection, where only a few annotated samples of novel classes are needed with abundant samples of base classes. 
To this end, we propose Prototypical VoteNet to recognize and localize novel instances, which incorporates two new modules: Prototypical Vote Module (PVM) and Prototypical Head Module (PHM). 
Specifically, as the 3D basic geometric structures can be shared among categories, PVM is designed to leverage class-agnostic geometric prototypes, which are learned from base classes, to refine local features of novel categories.
Then PHM is proposed to utilize class prototypes to enhance the global feature of each object, facilitating subsequent object localization and classification, which is trained by the episodic training strategy. 
To evaluate the model in this new setting, we contribute two new benchmark datasets, FS-ScanNet and FS-SUNRGBD.
We conduct extensive experiments to demonstrate the effectiveness of Prototypical VoteNet, and our proposed method shows significant and consistent improvements compared to baselines on two benchmark datasets.
This project will be available at \url{https://shizhen-zhao.github.io/FS3D_page/}.          

\end{abstract}

\section{Introduction}
3D object detection aims to localize and recognize objects from point clouds with many applications in augmented reality, autonomous driving, and robotics manipulation.
Recently, a number of fully supervised 3D object detection approaches have made remarkable progress with deep learning~\cite{Misra_2021_ICCV,Liu_2021_ICCV,Shi_2019_CVPR,Qi_2019_ICCV}.
Nonetheless, their success heavily relies on large amounts of labeled training data, which are time-consuming and costly to obtain.
On the contrary, a human can quickly learn to recognize novel classes by seeing only a few samples. 
To imitate such human ability, we consider few-shot point cloud 3D object detection, which aims to train a model to recognize novel categorizes from limited annotated samples of novel classes together with sufficient annotated data of base classes.

% \xjqi{please check and revise: train a model to recognize novel categorizes from limited annotated samples of novel classes together with sufficient annotate data of known base classes. 
% %This process is facilitated by known base categories with sufficient training samples to provide prior knowledge. 
% }
% train a model with limited annotations of novel classes and sufficient data for base classes in point cloud scenes \xjqi{please revise the sentence by highlighting what the base classes data are used for. This is too shallow. Check some few-shot papers for their insights. Do they refer to knowledge transfer and so on??}. 

Few-shot learning has been extensively studied in various 2D visual understanding tasks such as object detection~\cite{Wang_2020_ICML,Wang_2019_ICCV,Wu_2021_NIPS,Yan_2019_ICCV}, image classification~\cite{Kang_2019_ICCV,Finn_2017_ICML,Chen_2019_ICLR,Snell_17_NIPS}, and semantic segmentation~\cite{Nguyen_2019_ICCV,Min_2021_ICCV,Zhang_2021_CVPR,Lu_2021_ICCV}.
Early attempts~\cite{Finn_2017_ICML,Lee_2019_CVPR,He_2017_ICCV,Vinyals_16_NIPS} employ meta-learning to learn transferable knowledge from a collection of tasks and attained remarkable progress. 
Recently, benefited from large-scale datasets (\textit{e.g.} ImageNet~\cite{imagenet}) and advanced pre-training methods~\cite{clip,Tip_Adapter,Clip_adapter,cocoop}, finetuning large-scale pre-trained visual models on down-stream few-shot datasets emerges as an effective approach to address this problem~\cite{Song_2022_ACL,Wang_2020_ICML,Perceiver}. 
Among different streams of work, prototype-based methods~\cite{Wu_2021_ICCV,Zhu_2021_CVPR,Ma_2021_ICCV,Rectification} have been incorporated into both streams and show the great advantages, since they can capture the representative features of categories that can be further utilized for feature refinement~\cite{Yan_2019_ICCV,Zhao_2021_CVPR} or classification~\cite{Qiao_2021_ICCV,Snell_17_NIPS}. 

%However, the 3D point cloud community lacks large-scale datasets for pretraining, so that 3D detectors have to be trained from scratch. Consequently, the model can easily overfit base classes with poor generalization ability on novel classes, rendering the pre-training and fine-tuning scheme not a readily feasible solution to 3D few-shot learning. 
%
%This inspires us to rethink if meta learning is more effective in few-shot 3D point cloud object detection. 

% \xjqi{This inspires us to rethink previous few-shot xxx as xxxx. [relate it with the nice properties of protype based methods] }
%Recently, researchers find out that the simple pre-training and fine-tuning framework~\cite{Song_2022_ACL,Wang_2020_ICML} can well address the 2D few-shot learning problem.
%
%The success largely relies on large-scale pretraining on ImageNet or vision-language datasets.
%
%In contrast, the 3D point cloud community lacks large-scale datasets for pretraining, so that 3D detectors have to be trained from scratch.
%
%Consequently, they are more likely to overfit base classes, resulting in poor knowledge transfer ability to novel classes.

\begin{figure*}[!t]
\centering
\includegraphics[width=1\textwidth]{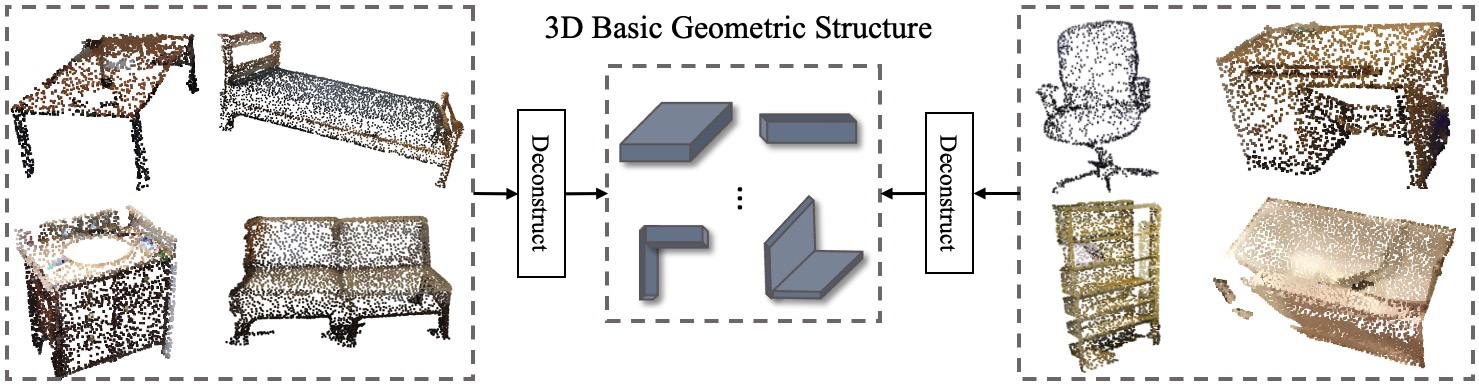}
% \vspace{-0.1in}
\caption{Illustration of the basic geometry of 3D objects, which can be shared among classes.
}
\label{fig:motivation}
\end{figure*}

This motivates us to explore effective 3D cues to build prototypes for few-shot 3D detection. 
Different from 2D visual data, 3D data can get rid of distortions caused by perspective projections, and offer geometric cues with accurate shape and scale information.
Besides, 3D primitives to constitute objects can often be shared among different categories.
For instance, as shown in Figure~\ref{fig:motivation}, rectangular plates and corners can be found in many categories. 
Based on these observations, in this work, we propose Prototypical VoteNet, which employs such robust 3D shape and primitive clues to design geometric prototypes to facilitate representation learning in the few-shot setting.
%This inspires us to design new 3D prototypes to leverage  such robust 3D shape and primitive clues to transfer robust knowledge of base categories to novel classes, facilitating feature learning in the few-shot setting. 

%inspires us that 3D basic geometric features can be utilize as a bridge to transfer robust knowledge of base categories to novel classes, facilitating the feature refinement in local geometric level.

% \xjqi{figure xxxx}. For instance, xxxx. 
% \xjqi{xxxx} can be used as a bridge to transfer robust knowledge of base categories to facilitate learning of novel categories.
% \xjqi{may revise later, I tried my best, but still have no cues to relate this with what we have done. Please also think of this , how the proposed module is related to the motivation here.}

%What's more, different from 2D visual scenes, the 3D cues of point clouds are more stable since it gets rid of some visual distractors, such as lighting and perspective, and can provide more accurate information of scale and shape.
%
%Besides, the 3D primitives to constitute objects can be shared among different categories.
%
%This offers new opportunities  to build prototypes leveraging \xjqi{xxx and xxx} for few-shot 3D point cloud object detection.

%In this work, we propose Prototypical VoteNet to address the few-shot 3D point cloud object detection task, which effectively incorporates 3D basic geometric prototypes and the meta-learning strategy for novel class feature refinement on local level and object level, respectively.
%
Prototypical VoteNet incorporates two new modules, namely Prototypical Vote Module (PVM) and Prototypical Head Module (PHM), to enhance local and global feature learning, respectively, for few-shot 3D detection.
Specifically, based on extracted features from a backbone network (\textit{i.e.} PointNet++~\cite{Qi_2017_NIPS}), PVM firstly constructs a class-agnostic 3D primitive memory bank to store geometric prototypes, which are shared by all categories and updated iteratively during training. 
To exploit the transferability of geometric structures, PVM then incorporates a multi-head cross-attention module to associate geometric prototypes with points in a given scene and utilize them to refine their feature representations.
PVM is majorly developed to exploit shared geometric structures among base and novel categories to enhance feature learning of local information in the few-shot setting. 
Further, to facilitate learning discriminative features for object categorization, PHM is designed to employ a multi-head cross-attention module and leverage class-specific prototypes from a few support samples to refine global representations of objects.
%
% A multi-head attention network is employed to utilize class-specific prototypes to refine object features in a given scene. 
%
Moreover, episodic training~\cite{Snell_17_NIPS,Vinyals_16_NIPS} is adopted to simulate few-shot circumstances, where PHM is trained by a distribution of similar few-shot tasks instead of only one target object detection task. 
%
% Given the consistency between the training few-shot task and the testing task, PHM is more generalizable to novel categories. \xjqi{Is this true??}
%
% In this way, supervised by the standard cross-entropy loss, PHM reduces the intra-class variation for the novel categories, with only a few support samples.

% Further, to facilitate learning discriminative features for object categorization, PHM is de

Our {\bf contributions} are listed as follows:
\begin{itemize}
\item We are the first to study the promising few-shot 3D point cloud object detection task, which allows a model to detect new classes, given a few examples.
\item We propose Prototypical VoteNet, which incorporates Prototypical Vote Module and Prototypical Head Module, to address this new challenge. 
Prototypical Vote Module leverages class-agnostic geometric prototypes to enhance the local features of novel samples.
Prototypical Head Module utilizes the class-specific prototypes to refine the object features with the aid of episode training.
\item  We contribute two new benchmark dataset settings called FS-ScanNet and FS-SUNRGBD, which are specifically designed for this problem. 
Our experimental results on these two benchmark datasets show that the proposed model effectively addresses the few-shot 3D point cloud object detection problem, yielding significant improvement over several competitive baseline approaches.
\end{itemize}

\section{Related Work}

\textbf{3D Point Cloud Object Detection.} 
Current 3D point cloud object detection approaches can be divided into two streams: Grid Projection/Voxelization based~\cite{Second,Lang_2019_CVPR,Song_2016_CVPR,Chen_2017_CVPR,Zhou_2018_CVPR,Wang_2020_ECCV} and point-based~\cite{Shi_2020_CVPR,Misra_2021_ICCV,Liu_2021_ICCV,Engelmann_2020_CVPR,Chen_2020_CVPR}. 
The former projects point cloud to 2D grids or 3D voxels so that the advanced convolutional networks can be directly applied.
%
% For example, Pixor~\cite{Yang_2018_CVPR} projects point cloud to the bird’s view and then employ 2D ConvNets for learning features and generating 3D boxes.
%
The latter methods take the raw point cloud feature extraction network such as PointNet++~\cite{Qi_2017_NIPS} to generate point-wise features for the subsequent detection. 
%
% Point RCNN~\cite{Shi_2019_CVPR} proposes to use a two-stage network, which is similar to Faster RCNN\cite{Ren_2015_NIPS}, to localize 3D objects.
% %
% VoteNet~\cite{Qi_2019_ICCV} introduces deep hough voting to accurately gather point cloud inside objects for object-level detection. 
%
Although these fully supervised approaches achieved promising 3D detection performance, their requirement for large amounts of training data precludes their application in many real-world scenarios where training data is costly or hard to acquire. 
To alleviate this limitation, we explore the direction of few-shot 3D object detection in this paper.

\textbf{Few-Shot Recognition.}
Few-shot recognition aims to classify novel instances with abundant base samples and a few novel samples.
%
% Hallucination-based methods~\cite{Hariharan_2017_ICCV,Wang_2018_CVPR} leverage hallucination techniques to augment the samples of novel categories.
%
Simple pre-training and finetuning approaches first train the model on the base classes, then finetune the model on the novel categories~\cite{Chen_2019_ICLR,Dhillon_2019_arxiv}.  
Meta-learning based methods~\cite{Finn_2017_ICML,Lee_2019_CVPR,He_2017_ICCV,Vinyals_16_NIPS,Snell_17_NIPS} are proposed to learn classifier across tasks and then transfer to the few-shot classification task. 
The most related work is Prototypical Network~\cite{Snell_17_NIPS}, which represents a class as one prototype so that classification can be performed by computing distances to the prototype representation of each class.
The above works mainly focus on 2D image understanding.
Recently, some few-shot learning approaches for point cloud understanding~\cite{Sharma_2020_NIPS,Zhao_2021_CVPR,Ye_2022_WACV} are proposed.
For instance, Sharma \textit{et al.}~\cite{Zhao_2021_CVPR} propose a graph-based method to propagate the knowledge from few-shot samples to the input point cloud. 
However, there is no work studying few-shot 3D point cloud object detection.
In this paper, we first study this problem and introduce the spirit of Prototypical Network into few-shot 3D object detection with 3D geometric prototypes and 3D class-specific prototypes.

\textbf{2D Few-shot Object Detection.} 
Most existing 2D few-shot detectors employ a meta-learning~\cite{Wang_2019_ICCV,Kang_2019_ICCV,Yan_2019_ICCV} or fine-tuning based mechanism~\cite{Wu_2020_ECCV,Wu_2021_NIPS,Qiao_2021_ICCV,Sun_2021_CVPR}.
Particularly, Kang \textit{et al.}~\cite{Kang_2019_ICCV} propose a one-stage few-shot detector which contains a meta feature learner and a feature re-weighting module. 
Meta R-CNN~\cite{Yan_2019_ICCV} presents meta-learning over RoI (Region-of-Interest) features and incorporates it into Faster R-CNN~\cite{Ren_2015_NIPS} and Mask R-CNN~\cite{He_2017_ICCV}. 
TFA~\cite{Wang_2020_ICML} reveals that simply fine-tuning the box classifier and regressor outperforms many meta-learning based methods.
Cao \textit{et al.}~\cite{Cao_2021_NIPS} improve the few-shot detection performance by associating each novel class with a well-trained base class based on their semantic similarity.
%
% {In this paper, we incorporate the meta-learning approach from Meta R-CNN~\cite{Yan_2019_ICCV} and the finetuning based method TFA~\cite{Wang_2020_ICML} into our benchmark, as they attained remarkable progress in 2D few-shot detection.} 
%
% In this paper, we choose Meta R-CNN~\cite{Yan_2019_ICCV} and TFA~\cite{Wang_2020_ICML} as the baseline method, due to their  by incorporating them to the 3D object detector 
% \xjqi{relate  this with your work and if there any method we have included in the benchmark [due to its high performance??] please highlight}

\begin{figure*}[!t]
\centering
\includegraphics[width=\textwidth]{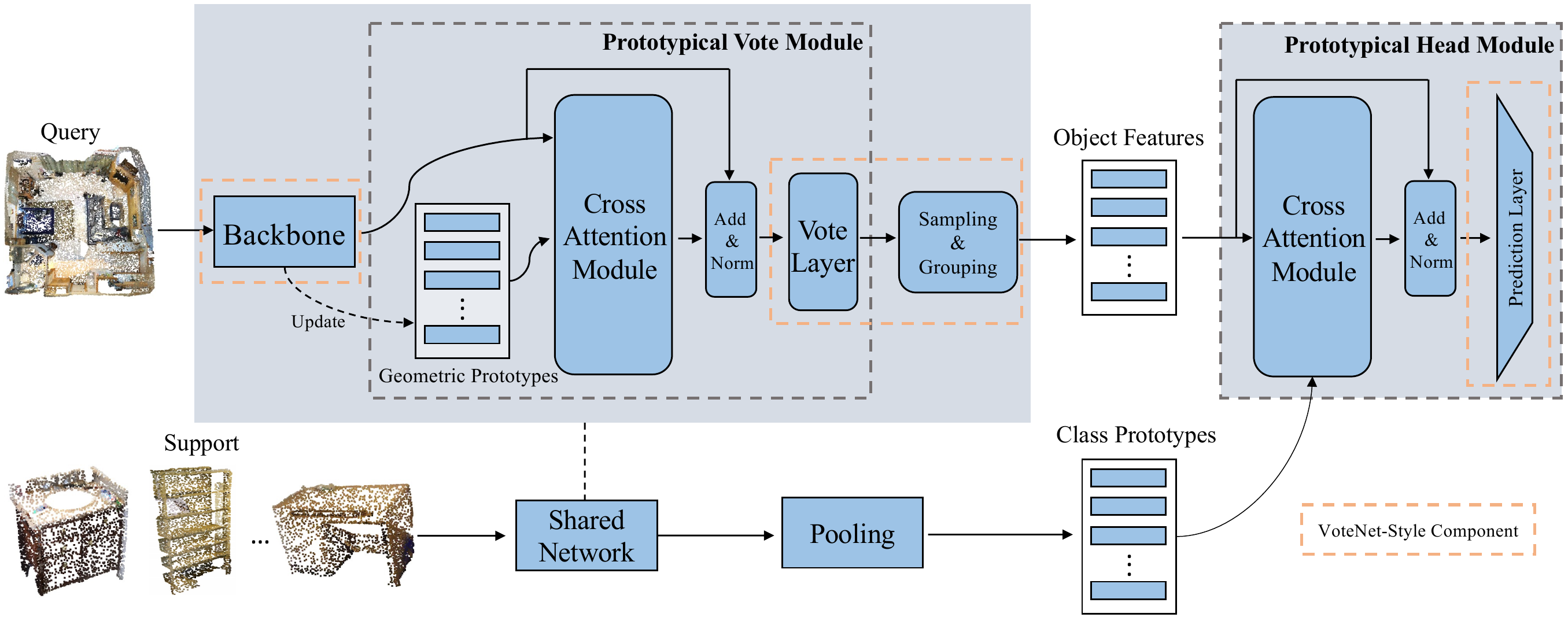}
\caption{Illustration of Prototypical VoteNet.
Prototypical VoteNet introduces two modules for few-shot 3D detection: 
1) Prototypical Vote Module for enhancing local feature representation of novel samples by leveraging the geometric prototypes,
2) Prototypical Head Module for refining global features of novel objects, utilizing the class-specific prototypes. 
}
\label{fig:approach}
\end{figure*}

\section{Our Approach}

%\subsection{Problem Definition}

In few-shot 3D point cloud object detection (FS3D), the object class set  $\mathbb C$ is split into $\mathbb C_\text{base}$ and $\mathbb C_\text{novel}$ such that $\mathbb C = \mathbb C_\text{base} \cup \mathbb C_\text{novel}$ and $\mathbb C_\text{base} \cap \mathbb C_\text{novel} = \emptyset$. 
For each class $r \in \mathbb C$, its annotation dataset $T_r$ contains all the data samples with object bounding boxes, that is $T_r = \{(u, P) | u \in \mathbb{R}^6, P \in \mathbb{R}^{N\times3}\}$. Here, $(u,P)$ is a 3D object bounding box $u = (x,y,z,h,w,l)$, representing box center locations and box dimensions, in a point cloud scene $P$.

%
%For each novel class $c \in \mathbb C_\text{novel}$, 
% There are only $K$ (\textit{e.g.} $K = 1$) examples/shots for each novel class $r$ ($|T_r| = K$), which are also known as support samples. 

There are only a few examples/shots for each novel class $r \in \mathbb{C}_\text{novel}$, which are known as support samples. 
Besides, there are plenty of annotated samples for each base class $r \in \mathbb{C}_\text{base}$.
Given the above dataset, FS3D aims to train a model to detect object instances in the novel classes leveraging such sufficient annotations for base categories $\mathbb C_\text{base}$ and limited annotations for novel categories $\mathbb C_\text{novel}$.
%with few annotated object instances. 
%
%Formally, given a query point cloud $P_q \in \mathbb{R}^{N*3}$, a few-shot point cloud detector is expected to output a set of detections $S_q = \{(c, u) | c \in C_{novel}, u \in U\}$.

In the following, we introduce Prototypical VoteNet for few-shot 3D object detection. 
We will describe the preliminaries of our framework in Section~\ref{sec:VoteNet}, which adopts the architecture of VoteNet-style 3D detectors~\cite{Qi_2019_ICCV,H3DNet,Chen_2020_CVPR}.
Then, we present Prototypical VoteNet consisting of Prototypical Vote Module (Section~\ref{sec:pvm}) and Prototypical Head Module (Section~\ref{sec:phm}) to enhance feature learning for FS3D.

\subsection{Preliminaries}
\label{sec:VoteNet}

VoteNet-style 3D detectors~\cite{Qi_2019_ICCV,H3DNet,Chen_2020_CVPR} takes a point cloud scene $P_i$ as input, and localizes and categorizes 3D objects. 
% Taking a point cloud scene $P_i$ as input, VoteNet~\cite{Qi_2019_ICCV} is to localize and categorize 3D objects.
As shown in Figure~\ref{fig:approach}, it firstly incorporates a 3D backbone network (\textit{i.e.} PointNet++~\cite{Qi_2017_NIPS}) parameterized by $\theta_1$ with downsampling layers for point feature extraction as Equation~\eqref{eq:equation}.
\begin{equation}
	F_i = h_1(P_i;\theta_1),
	\label{eq:equation}
\end{equation}
where $N$ and $M$ represent the original and subsampled number of points, respectively, $P_i \in \mathbb{R}^{N\times 3}$ represents an input point cloud scene $i$, and $F_i \in \mathbb{R}^{M\times (3+d)}$ is the subsampled scene points (also called seeds) with $d$-dimensional features and $3$-dimensional location coordinates.
%The subsampled $M$ points are also called seeds.
%incorporates a backbone network (\textit{i.e.} PointNet++ \xjqi{add citation})

%VoteNet~\cite{Qi_2019_ICCV} is a feed-forward network that consumes a 3D point cloud and outputs object proposals for 3D object detection. 
%
%Specifically, it is comprised of a point cloud feature extraction module that enriches a subsampled set of scene points (called seeds) with high-dimensional features (from N×3 input points to $M×(3+d)$ seeds), given by, 
%
%\begin{equation}
%	F_i = h_1(P_i;\theta_1),
%	\label{eq:equation}
%\end{equation}
%
%where $P_i \in \mathbb{R}^{N\times 3}, F_i \in \mathbb{R}^{M\times (3+d)}$.
%
Then, $F_i$ is fed into the vote module with parameters $\theta_2$ which outputs a $3$-dimensional coordinate offset $\Delta d_j =(\Delta x_j, \Delta y_j, \Delta z_j)$ relative to its corresponding object center $c= (c_x, c_y, c_z)$ and a residual feature vector $\Delta f_j$ for each point $j$ in $F_i = \{f_j\}_i$ as in Equation~\eqref{eq:vote}.
\begin{equation}
	\{\Delta d_j, \Delta f_j\}_i = h_2(F_i;\theta_2).
	\label{eq:vote}
\end{equation}
%where $\Delta d_j=(\Delta x_j, \Delta y_j, \Delta z_j)$ is the Euclidean space offset  point $j$. 
%
Given the predicted offset $\Delta d_j$, the estimated corresponding object center $c_j = (c_{x_j},c_{y_j}, c_{z_j})$ that point $j$ belongs to  can be calculated as Equation~\eqref{eq:center}. 
\begin{equation}
    c_{x_j} = x_j + \Delta x_j, c_{y_j} = y_j + \Delta y_j, c_{z_j} = z_j + \Delta z_j. 
    \label{eq:center}
\end{equation}
Similarly, the point features are updated as $F_i \gets F_i+\Delta F_i$ where $\Delta F_i = \{\Delta f_j\}_i$.

%The predicted center $(c_{x_j}, c_{y_j}, c_{z_j})$ of an object to which point $j$ belongs can be expressed as: $c_{x_j} = x_j+\Delta x_j, c_{y_j} = y_j +\Delta y_j, c_{z_j} = z_j+\Delta z_j$, 
{
Next, the detector samples object centers from $\{(c_{x_j}, c_{y_j}, c_{z_j})\}_i$ using farthest point sampling and group points with nearby centers together
%The estimated point centers $\{(c_{x_j}, c_{y_j}, c_{z_j})\}_i$ for all points in a scene $i$ are further used to group points with similar centers together
} (see Figure~\ref{fig:approach}: Sampling $\&$ grouping) to form a set of object proposals $O_i = \{o_t\}_i$. 
Each object proposal is characterized by a feature vector $f_{o_t}$ which is obtained by applying a max pooling operation on features of all points belonging to $o_t$.
%where $o_t$ represents an object proposal consisting of points belonging to it.

Further, equipped with object features $\{f_{o_t}\}_i$, the prediction layer with parameters $\theta_3$ is adopted to yield the bounding boxes $b_t$, objectiveness scores $s_t$, and classification logits $r_t$  for each object proposal $o_t$ following Equation~\eqref{eq:predict}.
\begin{equation}
	\{b_t, s_t, r_t\}_i = h_3(\{f_{o_t}\}_i;\theta_3).
	\label{eq:predict}
\end{equation}

% Finally, $\{b_t, s_t, r_t\}_i$ is processed with to produce the final 3D object detection result.

% the point features are updated as $F_i \gets F_i+\Delta F_i$.
%
%After the vote module, VoteNet samples points among the predicted centers, and aggregate their nearby points to generate object features $O_i=\{o_t\}_i$, where vector $o_t$ represents the feature of $t$-th object in a given point cloud $P_i$. 
%
%Then the object features processed by another point cloud network with parameters $\theta_3$ to predict 3D bounding boxes, objectness scores, and classification scores, 
%
%\begin{equation}
%	\{b_t, c_t, s_t\}_i = h_3(O_i;\theta_3),
%	\label{eq:predict}
%\end{equation}
%
%where $b_t$ is the predicted bounding box, $c_t$ is the predicted category, $s_i$ is the objectness score. 
%

%This process is equivalent to the pipeline in Fig.~\ref{fig:approach} without the two cross-attention modules,the branch for support point cloud, geometric prototypes, and class prototypes.  

% 
\subsection{Prototypical VoteNet}

Here, we present Prototypical VoteNet which incorporates two new designs -- Prototypical Vote Module (PVM) and Prototypical Head Module (PHM) to improve feature learning for novel categories with few annotated samples (see Figure~\ref{fig:approach}).
Specifically, PVM builds a class-agnostic memory bank of geometric prototypes $\mathcal{G}=\{g_k\}_{k=1}^K$ with a size of $K$, which models transferable class-agnostic 3D primitives learned from rich base categories, and further employs them to enhance local feature representation for novel categories via a multi-head cross-attention module. The enhanced features are then utilized by the Vote Layer to output the offset of coordinates and features as Equation~\eqref{eq:vote}.
%\begin{equation}
%	\{\Delta d_j, \Delta f_j\}_i = h_2^p(F_i, \mathcal{G};\theta_2^p),
%	\label{eq:protovote}
%\end{equation}
%where $h_2^p$ represents the cross-attention module together with the VoteMLP with parameters $\theta_2^p$.
%to leverage class-agnostic 3D primitives from base categories and class-specific object representations from novel categories  
%As shown in Figure~\ref{fig:approach}, toward 3D few-shot object detection, we 
%propose Prototypical VoteNet, which incorporates two prototype guided modules on VoteNet: PVM and PHM. 
%
%First, as shown in Figure~\ref{fig:motivation}, since the basic geometry to constitute objects can be shared among categories, 3D primitives can be utilized as a bridge to transfer the knowledge of base classes to novel ones. 
%
%Inspired by this, PVM is designed to leverage the class-agnostic geometric prototypes $\mathcal{G}=\{g_k\}_{k=1}^K$, which is learned from base categories, to enhance the local features of novel classes. 
%
%Therefore, with parameter $\theta_2^p$, Equation~\ref{eq:vote} can be re-formuated as, 
%
%\begin{equation}
%	\{\Delta d_j, \Delta f_j\}_i = h_2^p(F_i, \mathcal{G};\theta_2^p),
%	\label{eq:protovote}
%\end{equation}
%
Second, to facilitate learning discriminative features for novel class prediction, PHM employs an attention-based design to leverage class-specific prototypes $\mathcal{E}=\{e_r\}_{r=1}^{R}$ extracted from the support set $D_{\text {support}}$ with $R$ categories to refine global discriminate feature for representing each object proposal (see Figure~\ref{fig:approach}).
The output features are fed to the prediction layer for producing results as Equation~\eqref{eq:predict}. 
%Compared with Equation~\ref{eq:predict}, with parameter $\theta_3^p$, this can be re-formulated as, %
%\begin{equation}
%	\{b_t, s_t, r_t\}_i = h_3^p(O_i, \mathcal{E};\theta_3^p),
%	\label{eq:predict2}
%\end{equation}
%
%where $h_3^p$ indicates the multi-head attention module with the prediction layer, and $\theta_3^p$ are their parameters.
To make the model more generalizable to novel classes, we exploit the episodic training~\cite{Snell_17_NIPS,Vinyals_16_NIPS} strategy to train PHM, where a distribution of similar few-shot tasks instead of only one object detection task is learned in the training phase.
PVM and PHM are elaborated in the following sections. 

\begin{figure*}[!t]
\centering
\includegraphics[width=1\textwidth]{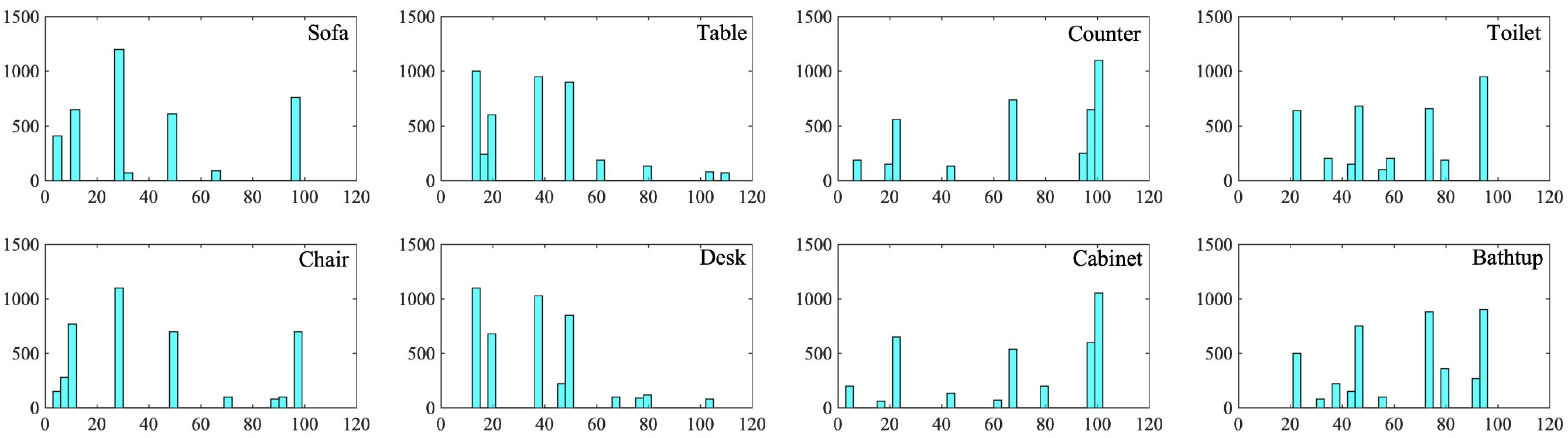}
% \vspace{-0.1in}
\caption{Visualization of the assignment of object point features to the geometric prototypes, where the horizontal axis represents the index of geometric prototypes and the vertical axis represents the number of assignments. 
}
\label{fig:histo}
\end{figure*}

\subsubsection{Prototypical Vote Module}

\label{sec:pvm}

Given input features $F_i$ extracted by a backbone network, Prototypical Vote Module constructs class-agnostic geometric prototypes $\mathcal{G} =\{g_k\}_{k=1}^K$ and then uses them to enhance local point features with an attention module.

\paragraph{Geometric Prototype Construction.} 
At the beginning, $\mathcal{G} = \{g_k\}_{k=1}^K \in \mathbb R^{d\times K}$ is randomly initialized.
During training, $\mathcal{G}$ is iteratively updated with a momentum based on point features of foreground objects.    
%After the above feature refinement step, PVM updates the memory bank of geometric prototypes $\mathcal{G}$ using the point features of foreground objects. 
%
Specifically, for each update, given $\mathcal{G}=\{g_k\}_{k=1}^K $ and all the foreground points $\{p_m\}_{m=1}^{M_f}$ with features $\{f_m\}_{m=1}^{M_f}$, where $M_f$ is the number of foreground points in the current batch, we assign each point to its nearest geometric prototype based on feature space distance. Then, for each prototype $g_k$, we have a group of points $\{p_m\}_k$ with features represented as $\{f_m\}_k$ assigned to it. Point features in one group are averaged to update the corresponding geometric prototype as Equation~\eqref{eq:update}.

\begin{equation}
    g_{k} \gets  \gamma*g_{k} + (1-\gamma) \overline{f}_{k}, \text{where}~ \overline{f}_k = \text {average}(\{f_m\}_k). 
    \label{eq:update}
\end{equation}

%\begin{equation}
% \overline{p}_{k} = \frac{1}{D_k}\sum_{p_d \in \{p_{d_k}\}_{d_k=1}^{D_k}} p_{d_k}, \label{eq:centroid}
%\end{equation}

%we calculate the nearest prototypes $\mathcal{G}=\{g_k\}_{k=1}^K$ that each point $p_m$ belongs based on the Euclidean space their Euclidean distances,
%
%\begin{equation}
  %  \vspace{-1pt}
 %   g_k =  \mathop{\arg\max}_{g_k \in \mathcal{G} } \langle p_d, g_k \rangle ,
 %   \label{eq:step2}
 %       \vspace{-1pt}
%\end{equation}
%
%For each $g_k$, we can obtain a point group $\{p_{d_k}\}_{d_k=1}^{D_k}$, and all the nearest geometric prototypes of point features in this group are $g_k$. 
%
%Then we calculate the mean feature vector for each point group,
%
%\begin{equation}
% \overline{p}_{k} = \frac{1}{D_k}\sum_{p_d \in \{p_{d_k}\}_{d_k=1}^{D_k}} p_{d_k}, \label{eq:centroid}
%\end{equation}
%
%Finally, we update each $g_k$ by, 
%
%\begin{equation}
%\label{eq:m_ins}
%    g_{k} \gets  m*g_{k} + (1-m) \hat{p}_{k},
%    \label{eq:update}
%\end{equation}
%
Here $\gamma \in [0,1]$ is the momentum coefficient for updating geometric prototypes in a momentum manner, serving as a moving average over all training samples.
Since one point feature is related to one geometric prototype, we call this one-hot assignment strategy as hard assignment.
An alternative to the hard assignment is the soft assignment, which calculates the similarity between a point features with all geometric prototypes. 
Empirically, we found that hard assignment results in more effective grouping versus soft assignment.
More details can be found in the supplementary material.

% \xjqi{Please add more analysis why we design it like this. The hard assignment seems very naive please explain why adopt the hard assignment to do the update, and what this used for. why it captures local information and the properties of this prototype add some explanation. We may move the visualization here if needed}

%$\hat{p}_{k}$ is $l_2$-normalized $\overline{p}_{k}$.
%

\paragraph{Geometric Prototypical Guided Local Feature Enrichment.} 
%
% {\xjqi{Given xxx, xxx, and xxx, the PVM module employs multi-head attention module to xxx as Equation xxx. } }
Given the geometric prototypes $\mathcal{G} = \{g_k\}_{k=1}^K$ and point features $F_i=\{f_j\}_i$ of a scene $i$, PVM further employs a multi-head cross-attention module~\cite{Transformer} to refine the point features. %as Equation~\ref{eq:vote_refine}. 
Specifically, the multi-head attention network uses the point features $F_i=\{f_j\}_i$ as query, geometric prototypes $\mathcal{G} = \{g_k\}_{k=1}^K$ as key and value where linear transformation functions with weights represented as $Q_h, U_h, V_h$ are applied to encode query, key and value respectively. Here, $h$ represents the head index. 
Then, for each head $h$, the query point feature is updated by softly aggregating the value features where the soft aggregation weight is determined by the similarity between the query point feature and corresponding key feature. The final point feature $f_j$ is updated by summing over outputs from all heads as Equation~\eqref{eq:vote_refine}.
% \xjqi{can think of drawing a figure for the multi-head attention}
%
\begin{equation}
f_j \leftarrow \text{Cross\_Att}(f_{j}, \{g_k\}) = \sum_{h=1}^{H} W_h  ( \sum_{k=1}^{K} A^{h}_{j,k} \cdot V_h g_k), \text{where}~ A^{h}_{j,k} = \frac{\exp[(Q_h f_{j})^T (U_h g_k)]}{\sum_{k=1}^{K}{\exp[(Q_h f_{j})^T (U_h g_k)]}}.
\label{eq:vote_refine}
\end{equation}
%
%\begin{equation}
%\label{eq:att_weight}
%A^{h}_{j,k} = \frac{\exp[(Q_h f_{j})^T %(U_h g_k)]}{\sum_{k=1}^{K}{\exp[(Q_h %f_{j})^T (U_h g_k)]}}
%\end{equation}
Here, $A_{j,k}^h$ is the soft aggregation weight considering the similarity between the $j$-th query point feature and the $k$-th key feature and used to weight the $k$-th value feature.
Through this process, the point feature is refined using geometric prototypes in a weighted manner where prototypes similar to the query point feature will have higher attention weights. This mechanism transfers geometric prototypes learned from base categories with abundant data to model novel points. The multi-head design enables the model to seek similarity measurements from different angles in a learnable manner to improve robustness.
Additionally, in both PHM and PVM, the multi-head attention layer are combined with feed forward FC layers.
After refining point features $\{f_j\}_i$, PVM predicts the point offset and residual feature vector $\{\Delta d_j, \Delta F_j\}_i$ as stated in Equation ~\eqref{eq:vote}. 
$\Delta d_j$ is explicitly supervised by a regression loss $\mathcal{L}_{\text{vote}}$ used in \cite{Qi_2017_NIPS}.

\paragraph{What do the geometric prototypes store?}

To shed more insights on what the geometric prototypes represent, we visualize the frequency of geometry prototypes in different categories using the ``assignment''.
The number of ``assignment''s of object point features to the geometric prototypes is shown in Figure~\ref{fig:histo}, where a point is assigned to the geometric prototype with the highest similarity.
In each histogram, the horizontal axis represents the index of geometric prototypes and the vertical axis represents the number of assignments. 
%
% We conduct this experiment in the testing set of our proposed benchmark datasets. 
%
Note that the first row is the novel classes and the second row is the base classes.
Figure~\ref{fig:histo} shows that two visually similar categories have a similar assignment histogram since they share the basic geometric structures.
This indicates that the memory bank of geometric prototypes successfully learns the 3D basic geometric knowledge, which can be a bridge to transfer the geometric knowledge from base classes to novel ones.

% To enhance local feature representations for novel classes, we design PVM, which builds up a class-agnostic 3D primitive memory bank to store geometric prototypes.
% %
% Our motivation is that  categories share the similar basic geometric shapes so that the geometric prototypes, which are mainly learned from base classes, can be transferred to novel classes. 
% %
% PVM input the geometric prototypes into a cross attention module, which is a multi-head attention network, to enhance the input point features. 
% %
% In the training stage, the forward process of PVM can be separated into two steps: 1) Update feature representation, 2) Update geometric prototypes.

%
% The predicted 3D offset $\Delta d_i$ is explicitly supervised by a regression loss $\mathcal{L}_{vote}$, which can be referred to VoteNet~\cite{Qi_2019_ICCV} for details. 

\subsubsection{Prototypical Head Module}
\label{sec:phm}
 
As shown in Figure~\ref{fig:approach}, given object proposals $O_i = \{o_t\}_i$ with features $\{f_{o_t}\}_i$ from Sampling \& Grouping module,
%$O_i = \{o_t\}_i$ with features $\{f_{o_t}\}_i$ obtained as detailed in \xjqi{Sec xxx} for a point cloud scene $i$
PHM module leverages class-specific prototypes $\{e_r\}$ to refine the object features $f_{o_t}$
%to enhance its effectiveness with an attention module 
for the subsequent object classification and localization.  
Moreover, for better generalizing to novel categories, PHM is trained by the episodic training scheme, where PHM learns a large number of similar few-shot tasks instead of only one task. 
Considering the function of PHM, we construct the few-shot tasks that, in each iteration, PHM refines the object features $\{f_{o_t}\}_i$ with the aid of class-specific prototypes, which are extracted from randomly sampled support samples. 
%
% The details are given as follows.

%the given samples for novel classes in the inference stage and
In each few-shot task, class-specific prototypes are built based on support sets that are randomly sampled. 
For class $r$, the class-specific prototype $e_r$ is obtained by averaging the instance features for all support samples in class $r$.
The instance feature is derived by applying a max pooling operation over the features of all points belonging to that instance.
As shown in Figure~\ref{fig:approach}, with class-specific prototypes $\mathcal{E}=\{e_r\}_{r=1}^{R}$ for a total of $R$ categories, PHM further employs a multi-head cross-attention module to refine object features $\{f_{o_t}\}_i$. %as Equation ~\ref{eq:phm}.
Here, the object features $\{f_{o_t}\}_i$ serve as the query features, class-specific prototypes are used to build value features and key features similar as what has been described in Section~\ref{sec:pvm}.
Then, the representation $f_{o_t}$ of each proposal $o_t$ is refined using the outputs of the multi-head attention module, which are weighted sum over the value features and the weight is proportionally to the similarity between the query feature and corresponding key features.
This process can be formulated as Equation~\eqref{eq:phm_refine}, which is similar to Equation~\eqref{eq:vote_refine}. 
\begin{equation}
f_{o_t} \leftarrow \text{Cross\_Att}(f_{o_t}, \{e_r\}).
\label{eq:phm_refine}
\end{equation}
Until now, $f_{o_t}$ is refined using class-specific prototypes, injecting class-specific features from given support samples into object-level features.
Finally, the refined object features $\{f_{o_t}\}_i$ are fed into the prediction module following Equation~\eqref{eq:predict}.

\begin{figure*}[!t]
\centering
\includegraphics[width=0.9\textwidth]{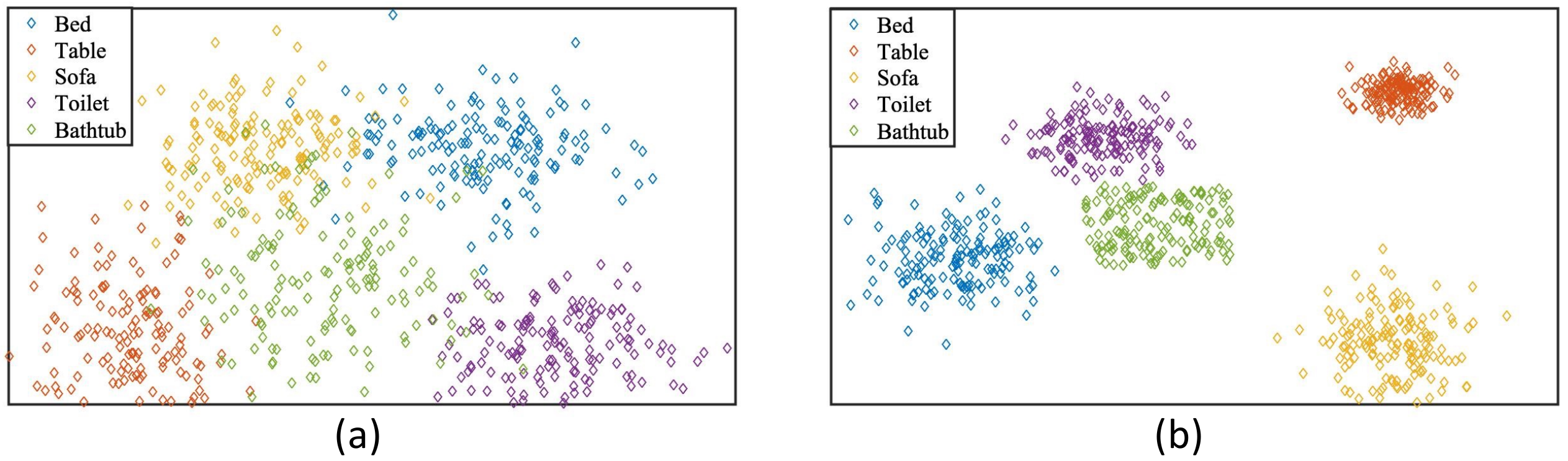}
% \vspace{-0.1in}
\caption{t-SNE visualization of the effect of PHM.
(a) shows the features without being processed by PHM.
(b) shows the features processed by PHM using class-specific prototypes.
}
\label{fig:tsne}
\end{figure*}

\paragraph{What does PHM help?}
Figure~\ref{fig:tsne} visualizes the effect of feature refinement by PHM.
The experiment is conducted on FS-ScanNet. 
Figure~\ref{fig:tsne}(a) shows the object features, which have not been processed by PHM. 
Figure~\ref{fig:tsne}(b) shows the object features processed by PHM. 
We could observe that, after the feature refinement by PHM, the features of each classes become more compact compared to the non-PHM one, which further validates the effectiveness of PHM.

\subsection{Model Training}
The model is trained by the episodic training strategy~\cite{Snell_17_NIPS,Vinyals_16_NIPS}.
The detailed training strategy is included in the supplementary material.  
We use the standard cross entropy loss $\mathcal{L}_\text{cls}$ and smooth-$L_1$ loss~\cite{Ren_2015_NIPS} $\mathcal{L}_\text{reg}$ to supervise the classification and the bounding box regression, respectively.
As for the objectness prediction, if a vote is located either within 0.3 meters to a ground truth object center or more than 0.6 meters from any center, it is considered to be positive~\cite{Qi_2019_ICCV}, which is supervised by a cross entropy loss $\mathcal{L}_\text{obj}$. 
Therefore, the overall loss for Prototypical VoteNet is given by, 
\begin{equation}
\mathcal{L}_\text{det} = \mathcal{L}_\text{cls} + \alpha_1 \mathcal{L}_\text{reg} + \alpha_2 \mathcal{L}_\text{obj} + \alpha_3 \mathcal{L}_\text{vote},
\end{equation}
where $\alpha_1, \alpha_2, \alpha_3$ is the coefficients to balance the loss contributions.

\section{Experiments} 

To our best knowledge, there is no prior study of few-shot point cloud object detection. 
Therefore, we setup a new benchmark which is described in Section~\ref{sec:Setup} \& ~\ref{sec:benchmark_method}.
Then, we conduct experiments and compare our method with baseline methods in Section~\ref{sec:main_results}.
Third, a series of ablation studies are performed for further analyzing Prototypical VoteNet in Section~\ref{sec:ablation}. 
In addition, the implementation details are included in the supplementary material.

\subsection{Benchmark Setup}

\label{sec:Setup}

\paragraph{Datasets.} 
We construct two new benchmark datasets FS-SUNRGBD and FS-ScanNet.
%
% We choose the indoor datasets because there are far more categories than outdoors where few-shot object detection is critical to enable models to recognize unseen c .
%
Specifically, \textbf{FS-SUNRGBD} is derived from SUNRGBD~\cite{Song_2015_CVPR}.
SUNRGBD consists of ~5K RGB-D training images annotated, and the standard evaluation protocol reports performance on 10 categories. 
We randomly select 4 classes as the novel ones while keeping the remaining ones as the base. 
In the training set, only $K$ annotated bounding boxes for each novel class are given, where k equals 1, 2, 3, 4 and 5. 
\textbf{FS-ScanNet} is derived from ScanNet~\cite{dai2017scannet}.
ScanNet consists of 1,513 point clouds, and the annotation of the point clouds corresponds to 18 semantic classes plus one for the unannotated space.
Out of its 18 object categories, we randomly select 6 classes as the novel ones, while keeping the remaining as the base. 
We evaluate with 2 different base/novel splits.
In the training set, only $K$ annotated bounding boxes for each novel class are given, where k equals 1, 3 and 5. 
More details about the new benchmark datasets can be referred to the supplementary material. 

\textbf{Evaluation Metrics.} 
We follow the standard evaluation protocol~\cite{Qi_2019_ICCV} in 3D point cloud object detection by using mean Average Precision(mAP) under different IoU thresholds (\textit{i.e.} 0.25, 0.50), denoted as AP$_{25}$ and AP$_{50}$. 
In addition, in the inference stage, we detect both novel classes and base classes. 
The performance on base classes is included in the supplementary material.

%

% \paragraph{Implementation Details.}
% %
% We follow recent practice~\cite{Qi_2019_ICCV,Xie_2020_CVPR} to use PointNet++~\cite{Qi_2017_NIPS} as a default backbone network. 
% %
% The backbone has 4 set abstraction layers and 2 feature propagation layers. 
% %
% For each set abstraction layer, the input point cloud is sub-sampled to 2048, 1024, 512, and 256 points with the increasing receptive radius of 0.2, 0.4, 0.8, and 1.2, respectively. 
% %
% Then, two feature propagation layers successively up-sample the points to 512 and 1024.
% %
% The size of memory bank of geometric prototypes is set to 120.
% %
% The number of head in both multi-head attention networks is empirically set to 4.
% %
% Momentum coefficient is set to 0.999.
% %
% We use 40k points in FS-ScanNet and 20k points in FS-SUNRGBD as input and adopt the same data augmentation as in ~\cite{Qi_2019_ICCV}, including a random flip, a random rotation between $\left[−5^\circ, 5^\circ\right]$, and a random scaling of the point cloud by [0.9, 1.1]. 
% %
% The network is trained from scratch by the AdamW optimizer with 36 epochs. 
% %
% The weight decay is set to 0.01. 
% %
% The initial learning rate is 0.008 in FS-ScanNet and 0.001 in FS-SunRGBD and decayed by 10× at the 24-th epoch and the 32-th epoch.
% %
% The gradnorm clip is applied to stabilize the training dynamics. Following~\cite{Qi_2019_ICCV} we use class aware head for box size prediction. 

\subsection{Benchmark Few-shot 3D Object Detection}

\label{sec:benchmark_method}

% \xjqi{We build the first benchmark for few-shot 3D object detection. The benchmark incorporates xxx baseline methods that have been shown to be successful in 2D few-shot object detection and our developed prototype votenet.} 
% We compare our model with five competitive baselines. 
%
We build the first benchmark for few-shot 3D object detection.
The benchmark incorporates 4 competitive methods, and three of them are built with few-shot learning strategies, which have been shown to be successful in 2D few-shot object detection.
\begin{itemize}
\item Baseline: We abandon PVM and PHM, and train the detector on the base and novel classes together. 
In this way, it can learn good features from the base classes that are applicable for detecting novel classes. 

\item VoteNet+TFA~\cite{Wang_2020_ICML}: We incorporate a well-designed few-shot object detection method TFA~\cite{Wang_2020_ICML} with VoteNet.
TFA first trains VoteNet on the training set with abundant samples of base classes.
Then only the classifier and the regressor are finetuned with a small balance set containing both base classes and novel classes. 
\item VoteNet+PT+TFA: The pretraining is proven to be important in 2D few-shot learning, as it learns more generic features, facilitating knowledge transfer from base classes to novel classes. 
Therefore, we add a pretraining stage, which is borrowed from a self-supervised point cloud contrastive learning method~\cite{Hou_2021_CVPR}, before the training stage of VoteNet+TFA. 
\item Meta VoteNet: 
Meta VoteNet is inspired by Meta RCNN~\cite{Yan_2019_ICCV}. 
Meta RCNN develops a meta-learning prediction head, which leverages class prototypes to multiply the RoI features and then predicts if a RoI feature is the category that the class prototype belongs to. 
In VoteNet, the grouped point features after the vote stage can be treated as the RoI features. 
\end{itemize}

\subsection{Main Results}

\label{sec:main_results}

\paragraph{FS-ScanNet.} 
We first compare the proposed approach with the baselines on FS-ScanNet in Table~\ref{tbl:scannet}.
These results show:
1) VoteNet+TFA has the worst performance.
For example, it only achieves 8.07\% AP$_{25}$ and 1.03\% AP$_{50}$ on 3-shot in Novel Split-1.
The reason is that VoteNet+TFA is trained from scratch, and only a few layers are funetuned by novel samples. 
Therefore, it tends to be overfitting with poor generalization ability on novel classes.
2) Even adding a pretaining stage to VoteNet+TFA, which is termed as VoteNet+PT+TFA, it still performs poorly. 
For example, it only contributes the improvement of +2.30\% AP$_{25}$ and +1.10\% AP$_{50}$ to VoteNet+TFA on 3-shot in Novel Split-1. 
This is different from 2D few-shot object detection.
This is potentially caused by the lack of large-scale datasets for pre-training in 3D, while the 2D backbone ResNet~\cite{He_2016_CVPR} has a dataset of over one million images for pre-training.
%
% Therefore, the pretraining stage in 3D can not facilitate VoteNet+PT+TFA to learn feature representation generic enough for transferring knowledge from base classes to novel classes. 
%
3) Besides our designed method, Meta VoteNet is superior. 
%
% For example, Meta VoteNet outperforms Baseline by +3.09\% AP$_{25}$ and +1.95\% AP$_{50}$ on 3-shot in Novel Split-1. 
For example, Meta VoteNet reaches 25.73\% AP$_{25}$ and 16.01\% AP$_{50}$ on 3-shot in Novel Split-1. 
This shows the meta-learning prediction head, derived from Meta RCNN~\cite{Yan_2019_ICCV}, helps to generalize from base classes to novel classes. 
4) Our new model Prototypical VoteNet surpasses all competitors by significant margins. 
For example, Prototypical VoteNet achieves 32.25\% AP$_{25}$ and 16.01\% AP$_{50}$ on 3-shot in Novel Split-1.
This is because our proposed method is equipped with a more generic vote module by learning geometric prototypes, and leverage class prototypes to promote the discriminative feature learning. 

\begin{table*}[t]
	\centering
	\scalebox{0.81}{
		\begin{tabular}{L{2.5cm}?C{0.6cm} C{0.6cm}? C{0.6cm} C{0.6cm}? C{0.6cm} C{0.6cm} ? C{0.6cm}  C{0.6cm}?  C{0.6cm}  C{0.6cm}? C{0.6cm} C{0.6cm}}
			\hline\toprule[0.1pt]
			\multirow{3}{*}{\textbf{Method}} 
			& \multicolumn{6}{c?}{\textbf{Novel Split 1}}
			& \multicolumn{6}{c}{\textbf{Novel Split 2}} \\ \cline{2-13} 
			& \multicolumn{2}{c?}{\textbf{1-shot}} &
			\multicolumn{2}{c?}{\textbf{3-shot}}&
			\multicolumn{2}{c?}{\textbf{5-shot}}
			& \multicolumn{2}{c?}{\textbf{1-shot}} &
			\multicolumn{2}{c?}{\textbf{3-shot}}
			&\multicolumn{2}{c}{\textbf{5-shot}}
			\\ \cline{2-13} 
			& \multicolumn{1}{c}{AP$_{25}$} & \multicolumn{1}{c?}{AP$_{50}$} &\multicolumn{1}{c}{AP$_{25}$} & \multicolumn{1}{c?}{AP$_{50}$} & \multicolumn{1}{c}{AP$_{25}$} & \multicolumn{1}{c?}{AP$_{50}$}  
			& \multicolumn{1}{c}{AP$_{25}$} & \multicolumn{1}{c?}{AP$_{50}$} &\multicolumn{1}{c}{AP$_{25}$} & \multicolumn{1}{c?}{AP$_{50}$} & \multicolumn{1}{c}{AP$_{25}$} & \multicolumn{1}{c}{AP$_{50}$}
			\\  \hline\toprule[0.1pt]
		    Baseline & 9.21 & 3.14  & 22.64  & 9.04  & 24.93  & 12.82 & 4.92  & 0.94 & 15.86  & 3.15  & 20.72  &  6.13 \\ \hline
			VoteNet+TFA  & 0.48 & 0.09  & 8.07 & 1.03  & 16.36  & 7.91  & 1.00 & 0.15  & 2.64  & 0.22  & 5.71  & 2.40\\ \hline
			VoteNet+PT+TFA & 2.58 & 1.04  & 10.37 & 2.13  & 17.21  & 8.94  & 2.13 & 0.56  & 4.85  & 1.25  & 7.25  & 2.49\\  \hline
			Meta VoteNet & 11.01 & 4.20  & 25.73 & 10.99  & 26.68  & 14.40  & 6.06 & 1.01  & 16.93  & 4.51  &23.83  & 7.17\\  \hline
			\toprule[0.1pt]
			\textbf{Ours}  & \textbf{15.34} & \textbf{8.25}  & \textbf{31.25}  & \textbf{16.01}  & \textbf{32.25}  & \textbf{19.52} & \textbf{11.01} & \textbf{2.21}  & \textbf{21.14}  &  \textbf{8.39} &  \textbf{28.52} & \textbf{12.35} \\ \hline\toprule[0.1pt]
		\end{tabular}
	}
	\caption{Results on \textbf{FS-ScanNet} using mean Average Precision (mAP) at two different IoU thresholds of 0.25 and 0.50, denoted as AP$_{25}$ and AP$_{50}$.}
% 	\caption{Results on \textbf{FS-ScanNet} using AP$_{25}$ and AP$_{50}$.}
	\label{tbl:scannet}
\end{table*}

\begin{table*}[t]
	\centering
	\scalebox{0.80}{
		\begin{tabular}{L{2.4cm}?C{0.6cm} C{0.6cm}? C{0.6cm} C{0.6cm}? C{0.6cm} C{0.6cm} ? C{0.6cm}  C{0.6cm}?  C{0.6cm}  C{0.6cm}}
			\hline\toprule[0.1pt]
			\multirow{2}{*}{\textbf{Method}} 
			& \multicolumn{2}{c?}{\textbf{1-shot}} &
			\multicolumn{2}{c?}{\textbf{2-shot}}&
			\multicolumn{2}{c?}{\textbf{3-shot}}
			& \multicolumn{2}{c?}{\textbf{4-shot}} &
			\multicolumn{2}{c}{\textbf{5-shot}}
			\\ \cline{2-11} 
			& \multicolumn{1}{c}{AP$_{25}$} & \multicolumn{1}{c?}{AP$_{50}$} &\multicolumn{1}{c}{AP$_{25}$} & \multicolumn{1}{c?}{AP$_{50}$} & \multicolumn{1}{c}{AP$_{25}$} & \multicolumn{1}{c?}{AP$_{50}$}  
			& \multicolumn{1}{c}{AP$_{25}$} & \multicolumn{1}{c?}{AP$_{50}$} &\multicolumn{1}{c}{AP$_{25}$} & \multicolumn{1}{c}{AP$_{50}$}
			\\  \hline\toprule[0.1pt]
		    Baseline & 5.46 &  0.22 & 6.52  & 0.77& 13.73  &  2.20 & 20.47 & 4.50 & 22.99  & 5.90   \\ \hline
			VoteNet+TFA  & 1.41 & 0.03 & 3.70  &  0.78 & 4.03  & 1.09 & 7.91 & 2.10  &  8.50 &  2.81  \\ \hline
			VoteNet+PT+TFA  & 3.40 & 0.51 & 5.13  & 1.22  & 7.94  & 2.31 & 10.05 & 3.12  & 11.32  & 4.01 \\ \hline
			Meta VoteNet & 7.04 &  0.98 & 9.23  & 1.34 & 16.24  &  3.12 & 20.10 & 4.69 & 24.41  & 6.05   \\ \hline 
			\textbf{Ours}  & \textbf{12.39} & \textbf{1.52}  & \textbf{14.54}  & \textbf{3.05}  & \textbf{21.51}  & \textbf{6.13} & \textbf{24.78}  & \textbf{7.17}  & \textbf{29.95}  &  \textbf{8.16}  \\ \hline\toprule[0.1pt]
		\end{tabular}
	}
	\caption{Results on \textbf{FS-SUNRGBD} using mean Average Precision (mAP) at two different IoU thresholds of 0.25 and 0.50, denoted as AP$_{25}$ and AP$_{50}$.}
% 	\caption{Results on \textbf{FS-SUNRGBD} using AP$_{25}$ and AP$_{50}$.}
	\label{tbl:sunrgbd}
\end{table*}

\paragraph{FS-SUNRGBD.} Table~\ref{tbl:sunrgbd} shows the result comparison on FS-SUNRGBD. 
In FS-SUNRGBD, we find that VoteNet+TFA and VoteNet+PT+TFA still have a large gap with other methods due to the overfitting and insufficient pretraining. 
Meanwhile, Meta VoteNet outperforms other baseline methods.
For example, on 3-shot, Meta VoteNet surpasses VoteNet+PT+TFA by +8.34\% AP$_{25}$ and +0.81\% AP$_{50}$ and achieves 16.24\% AP$_{25}$ and 3.12\% AP$_{50}$.
Finally, our designed Prototypical VoteNet outperforms all the methods. 
For instance, on 3-shot, Prototypical VoteNet surpasses Meta VoteNet by +5.37\% AP$_{25}$ and +3.01\% AP$_{50}$, which further validates the effectiveness of our proposed PVM and PHM.

\subsection{Further Analysis}

\label{sec:ablation}

\begin{minipage}[t]{\textwidth}
\begin{minipage}[]{0.50\textwidth}
\makeatletter\def\@captype{table}
% \begin{table*}[!t]
          \centering
    \begin{tabular}{@{}l|cc|cc@{}}
    \hline
    \multirow{2}{*}{\textbf{Method}}  & \multicolumn{2}{c?}{\textbf{3-shot}} & \multicolumn{2}{c}{\textbf{5-shot}}\\
    \cline{2-5}
    & AP$_{25}$ & AP$_{50}$ & AP$_{25}$ & AP$_{50}$ \\
    \hline
    Baseline & 22.64  & 9.04  & 24.93  & 12.82 \\
\hline
    +PVM & 27.43 & 13.63 & 28.44 & 16.45 \\
\hline
    +PHM & 28.76 & 14.04 & 30.13 & 17.51 \\
\hline
    +PVM+PHM  & \textbf{31.25}  & \textbf{16.01}  & \textbf{32.25}  & \textbf{19.52}\\
    \hline
                \end{tabular}
\caption{
% Ablation study on 3-shot and 5-shot in Split-1 of \textbf{FS-ScanNet} using mean Average Precision (mAP) at two different IoU thresholds of 0.25 and 0.50, denoted as AP$_{25}$ and AP$_{50}$.
Ablation study of individual components.}
\label{tbl:ablation}
% \end{table*}
\end{minipage}
\hfill
\begin{minipage}[]{0.48\textwidth}
\makeatletter\def\@captype{table}
 \centering
\begin{tabular}{l|l|cc}\hline
& Prototype & AP$_{0.25}$  & AP$_{0.50}$\\ \hline
\multirow{2}{*}{PVM}  &Geometric  & 31.25  & 16.01 \\
& Self-learning & 28.34 & 14.01 \\
\hline
\multirow{2}{*}{PHM} &Class  & 31.25  & 16.01 \\
& Self-learning  & 27.45 & 13.67 \\
\hline
\end{tabular}
\caption{Ablation study of Prototypes.}
\label{tbl:proto}
\end{minipage}
\end{minipage}

\paragraph{Contributions of individual components.}
We conduct ablation studies on 3-shot and 5-shot in split-1 of FS-ScanNet with results shown in Table~\ref{tbl:ablation}.
The results show that both of them are effective on their own. 
For example, on 3-shot, PVM contributes the improvement of +4.19\% AP$_{25}$ and +4.59\% AP$_{50}$, and PHM contributes the improvement of +6.12\% AP$_{25}$ and +5.00\% AP$_{50}$.
Moreover, when combined, the best performance, 31.25\% AP$_{25}$ and 16.01\% AP$_{50}$, is achieved. 

\paragraph{Effectiveness of Prototypes.} 
Table ~\ref{tbl:proto} shows the effectiveness of two kinds of prototypes.
In order to validate the geometric prototypes, we displace them by the self-learning ones, which are randomly initialized and updated by the gradient descend during the model training. 
Table ~\ref{tbl:proto} results show that the performance significantly degrades.
To validate the effectiveness of class prototypes, we also alter them by the randomly initialized self-learning prototypes. 
Similarly, the performance drops drastically due to the lack of class prototypes.

% \begin{table*}[!t]
%           \centering
%     \begin{tabular}{@{}l|cc|cc@{}}
%     \toprule
%     \multirow{2}{*}{\textbf{Method}}  & \multicolumn{2}{c?}{\textbf{3-shot}} & \multicolumn{2}{c}{\textbf{5-shot}}\\
%     \cline{2-5}
%     & AP$_{25}$ & AP$_{50}$ & AP$_{25}$ & AP$_{50}$ \\
%     \hline
%     Baseline & 22.64  & 9.04  & 24.93  & 12.82 \\
% \hline
%     +PVM & 27.43 & 13.63 & 28.44 & 16.45 \\
% \hline
%     +PHM & 28.76 & 14.04 & 30.13 & 17.51 \\
% \hline
%     +PVM+PHM  & \textbf{31.25}  & \textbf{16.01}  & \textbf{32.25}  & \textbf{19.52}\\
%     \hline\toprule[0.1pt]
%                 \end{tabular}
% \caption{
% Ablation study on 3-shot and 5-shot in Split-1 of \textbf{FS-ScanNet} using mean Average Precision (mAP) at two different IoU thresholds of 0.25 and 0.50, denoted as AP$_{25}$ and AP$_{50}$.}
% \label{tbl:ablation}
% \end{table*}

% \paragraph{Design choices in Geometric} 
% We validate the design choices in xxxx including: size of memory bank, coefficent xxx, distance measure xxx. \xjqi{please move corresponding part here.}

\begin{minipage}[t]{\textwidth}
\begin{minipage}[]{0.48\textwidth}
\makeatletter\def\@captype{table}
\centering
\begin{tabular}{l|cc}\hline
\# Prototype & AP$_{0.25}$  & AP$_{0.50}$\\ \hline
~~~$K=30$   & 29.98 & 15.01 \\
~~~$K=60$  & 30.24 & 15.54\\
~~~$K=90$  & 31.10 & 15.98\\
~~~$K=120$  & 31.25 & 16.01\\
~~~$K=150$  & 31.01  & 15.89 \\
\hline
\end{tabular}
\caption{Ablation study of memory bank size.}
\label{tbl:size}
\end{minipage}
\hfill
\begin{minipage}[h]{0.48\textwidth}
\makeatletter\def\@captype{table}
 \centering
\begin{tabular}{l|cc}\hline
Coefficient $m$ & AP$_{0.25}$  & AP$_{0.50}$\\ \hline
$\gamma = 0.2$ & 29.50  & 14.65 \\
$\gamma = 0.9$ & 30.55  & 15.15 \\
$\gamma = 0.99$  & 30.71  & 15.30 \\
$\gamma = 0.999$  & 31.01  & 15.89 \\
$\gamma = 0.9999$  &  30.90  & 15.79 \\
\hline
\end{tabular}
\caption{Ablation study of coefficient $\gamma$.}
\label{tbl:coefficient}
\end{minipage}
\end{minipage}

\paragraph{Size of Memory Bank.} 
Table~\ref{tbl:size} studies the size of the memory bank containing the geometric prototypes.
This ablation study is performed on 3-shot in split-1 of FS-ScanNet. 
The value of $\text{K}$ is set to $\{30, 60, 90, 120, 150\}$.
For $\text{K}=30$, the memory bank only contains 30 geometric prototypes, which only achieves 29.98\% AP$_{25}$ and 15.01\% AP$_{50}$.
Furthermore, when using more prototypes (i.e., $\text{K}=120$), there will be an obvious performance improvement, which reaches 31.25\% AP$_{25}$ and 16.01\% AP$_{50}$.
However, when continuous increasing $\text{K}$, there will be no improvement. 
Therefore, we set the size of the memory bank to be 120.

% \begin{figure}[h]
% \begin{minipage}[h]{0.48\textwidth}
% \makeatletter\def\@captype{table}
%  \centering
% \begin{tabular}{|l|c|c|}\hline
% Coefficient $m$ & AP$_{0.25}$  & AP$_{0.50}$\\ \hline\hline
% $\gamma = 0.2$ & 29.50  & 14.65 \\
% $\gamma = 0.9$ & 30.55  & 15.15 \\
% $\gamma = 0.99$  & 30.71  & 15.30 \\
% $\gamma = 0.999$  & 31.01  & 15.89 \\
% $\gamma = 0.9999$  &  30.90  & 15.79 \\
% \hline
% \end{tabular}
% \caption{Ablation study of coefficient $\gamma$.}
% \label{tbl:coefficient}
% \end{minipage}
% \hfill
% \centering
% \begin{minipage}[h]{0.5\textwidth}
% \includegraphics[width=1.0\linewidth,height=0.13\textheight]{figure/tsne.jpg}
% \caption{\xjqi{should be refined} Tsne visualization of the effect of PHM.}
% \label{fig:tsne}
% \end{minipage}
% \hfill
% \end{figure}

\paragraph{Coefficient $\gamma$.} 
Table~\ref{tbl:coefficient} shows the effect of momentum coefficient ($\gamma$ in Equation~\eqref{eq:update}). 
The experiment is performed on 3-shot in split 1 of FS-ScanNet. 
The results show that, when using a relatively large coefficient (\textit{i.e.}, {$\gamma\in[0.999,0.9999]$}), the model performs well, compared with the model using a small momentum coefficient (\textit{i.e.}, {$\gamma\in[0.9,0.99]$}). 
Moreover, the performance drops when using a small value of {{$\gamma=0.2$}}.
The is potentially because a small momentum coefficient might bring about unstable prototype representation with rapid prototype updating. 

% \subsection{Another Baseline: VoteNet+TFA$^*$}

% \label{sec:baseline}

\begin{table*}[h]
	\centering
	\scalebox{0.81}{
		\begin{tabular}{L{2.5cm}?C{0.6cm} C{0.6cm}? C{0.6cm} C{0.6cm}? C{0.6cm} C{0.6cm} ? C{0.6cm}  C{0.6cm}?  C{0.6cm}  C{0.6cm}? C{0.6cm} C{0.6cm}}
			\hline\toprule[0.1pt]
			\multirow{3}{*}{\textbf{Method}} 
			& \multicolumn{6}{c?}{\textbf{Novel Split 1}}
			& \multicolumn{6}{c}{\textbf{Novel Split 2}} \\ \cline{2-13} 
			& \multicolumn{2}{c?}{\textbf{1-shot}} &
			\multicolumn{2}{c?}{\textbf{3-shot}}&
			\multicolumn{2}{c?}{\textbf{5-shot}}
			& \multicolumn{2}{c?}{\textbf{1-shot}} &
			\multicolumn{2}{c?}{\textbf{3-shot}}
			&\multicolumn{2}{c}{\textbf{5-shot}}
			\\ \cline{2-13} 
			& \multicolumn{1}{c}{AP$_{25}$} & \multicolumn{1}{c?}{AP$_{50}$} &\multicolumn{1}{c}{AP$_{25}$} & \multicolumn{1}{c?}{AP$_{50}$} & \multicolumn{1}{c}{AP$_{25}$} & \multicolumn{1}{c?}{AP$_{50}$}  
			& \multicolumn{1}{c}{AP$_{25}$} & \multicolumn{1}{c?}{AP$_{50}$} &\multicolumn{1}{c}{AP$_{25}$} & \multicolumn{1}{c?}{AP$_{50}$} & \multicolumn{1}{c}{AP$_{25}$} & \multicolumn{1}{c}{AP$_{50}$}
			\\  \hline\toprule[0.1pt]
			VoteNet+TFA  & 0.48 & 0.09  & 8.07 & 1.03  & 16.36  & 7.91  & 1.00 & 0.15  & 2.64  & 0.22  & 5.71  & 2.40\\ \hline
		    VoteNet+TFA$^*$ & 10.38 & 3.96  & 23.77  & 9.83  & 26.02  & 13.96 & 5.12  & 0.95 & 16.23  & 3.72  & 21.89  &  6.76 \\ \hline
			\toprule[0.1pt]
			\textbf{Ours}  & \textbf{15.34} & \textbf{8.25}  & \textbf{31.25}  & \textbf{16.01}  & \textbf{32.25}  & \textbf{19.52} & \textbf{11.01} & \textbf{2.21}  & \textbf{21.14}  &  \textbf{8.39} &  \textbf{28.52} & \textbf{12.35} \\ \hline\toprule[0.1pt]
		\end{tabular}
	}
	\caption{Results of VoteNet+TFA$^*$ on \textbf{FS-ScanNet} using mean Average Precision (mAP) at two different IoU thresholds of 0.25 and 0.50, denoted as AP$_{25}$ and AP$_{50}$.}
	\label{tbl:scannet_an}
\end{table*}

\begin{table*}[h]
	\centering
	{
		\begin{tabular}{@{}l|cc|cc@{}}
    \toprule
    \multirow{2}{*}{\textbf{Method}}  & \multicolumn{2}{c?}{\textbf{3-shot}} & \multicolumn{2}{c}{\textbf{5-shot}}\\
    \cline{2-5}
    & AP$_{25}$ & AP$_{50}$ & AP$_{25}$ & AP$_{50}$ \\
\hline
    GroupFree + DeFRCN & 25.22 & 10.90 & 26.42 & 14.01 \\
\hline
    GroupFree + FADI & 25.73 & 11.02 & 27.12 & 14.32\\
\hline
    3DETR + DeFRCN & 26.01 &	10.95 &	26.88 &	14.45\\
\hline
    3DETR + FADI & 26.24 &	11.12 &	26.93 &	15.22\\
\hline
    Ours & 31.25 & 16.01 & 32.25 & 19.52 \\
    \hline\toprule[0.1pt]
\end{tabular}}
    \caption{More Methods Borrowed From 2D Few-Shot Object detection}
	\label{tbl:more}
\end{table*}

\paragraph{More Analysis on TFA.} TFA~\cite{Wang_2020_ICML} is a simple yet effective method in few-shot 2D object detection. 
Specifically, TFA first trains a detector on the training set with abundant samples of base classes and then finetunes the classifier and the regressor on a small balanced set, which contains both base classes and novel classes.   
More details can be referred to in~\cite{Wang_2020_ICML}.
However, as shown in Table 1 and Table 2 in the main paper, VoteNet+TFA performs poorly in few-shot 3D object detection.
As discussed in Section 4.3 in the main paper, the reason is that a few layers are finetuned by the samples of novel classes.
A question arises here: why does TFA in few-shot 2D object detection only need to train the classifier and the regressor on the novel samples?
We speculate that the large-scale pre-training on ImageNet~\cite{imagenet} helps.
To overcome this problem, in the first stage of TFA, we train the VoteNet on the training set with both novel classes and base classes. 
Then, in the second stage, we use a small balanced set containing both novel classes and base classes to finetune the classifier and the regressor. 
We denote this new baseline as VoteNet+TFA$^*$.
As shown in Table~\ref{tbl:scannet_an},  VoteNet+TFA$^*$ boosts the performance significantly. 
For example, in 3-shot in split-1 of FS-ScanNet, VoteNet+TFA$^*$ can achieve 26.02\% AP$_{25}$ and 13.96\% AP$_{50}$, while VoteNet+TFA only reaches 16.36\% AP$_{25}$ and 7.91\% AP$_{50}$.

\paragraph{More Methods Borrowed From 2D Few-Shot Object detection.}
We combine two SOTA 2D few-shot object detection techniques (i.e. DeFRCN~\cite{Qiao_2021_ICCV}, FADI~\cite{Cao_2021_NIPS}) and two SOTA 3D detectors (i.e. GroupFree~\cite{Liu_2021_ICCV}, 3DETR~\cite{Misra_2021_ICCV}). These two few-shot techniques are plug-in-play modules and can be easily incorporated into the different detection architectures. 
We conducted this experiment on 3-shot and 5-shot in split-1 of FS-ScanNet. The results in Table~\ref{tbl:more} show that our method still surpasses these methods by a large margin.
This is potentially because, in the 2D domain, they often build their model upon a large-scale pre-trained model on ImageNet. However, in the 3D community, there does not exist a large-scale dataset for model pre-training, which requires future investigations.
Therefore, these 2D few-shot object detection techniques might not be directly transferable to the 3D domain.
For future works, we might resort to the pre-training models in the 2D domain to facilitate the few-shot generalization on 3D few-shot learning and how these techniques can be combined with our method.

\section{Concluding Remarks}
In this paper, we have presented Prototypical VoteNet for FS3D along with a new benchmark for evaluation. Prototypical VoteNet enjoys the advantages of two new designs, namely Prototypical Vote Module (PVM) and Prototypical Head Module (PHM), for enhancing feature learning in the few-shot setting.
Specifically, PVM exploits geometric prototypes learned from base categories to refine local features of novel categories. 
PHM is proposed to utilize class-specific prototypes to promote discriminativeness of object-level features. Extensive experiments on two new benchmark datasets demonstrate the superiority of our approach. 
We hope our studies on 3D propotypes and proposed new benchmark could inspire further investigations in few-shot 3D object detection.

\begin{ack}
This work has been supported by Hong Kong Research Grant Council - Early Career Scheme (Grant No. 27209621), HKU Startup Fund,  HKU Seed Fund for Basic Research, and Tencent Research Fund. Part of the described research work is conducted in the JC STEM Lab of Robotics for Soft Materials funded by The Hong Kong Jockey Club Charities Trust.
\end{ack}

\medskip

{
\bibliographystyle{splncs04}
\bibliography{reference}

\begin{thebibliography}{10}
\providecommand{\url}[1]{\texttt{#1}}
\providecommand{\urlprefix}{URL }
\providecommand{\doi}[1]{https://doi.org/#1}

\bibitem{Cao_2021_NIPS}
Cao, Y., Wang, J., Jin, Y., Wu, T., Chen, K., Liu, Z., Lin, D.: Few-shot object
  detection via association and discrimination. In: Advances in Neural
  Information Processing Systems (2021)

\bibitem{Chen_2020_CVPR}
Chen, J., Lei, B., Song, Q., Ying, H., Chen, D.Z., Wu, J.: A hierarchical graph
  network for 3d object detection on point clouds. In: IEEE Conference on
  Computer Vision and Pattern Recognition (2020)

\bibitem{Chen_2019_ICLR}
Chen, W.Y., Liu, Y.C., Kira, Z., Wang, Y.C.F., Jia-Bin, H.: A closer look at
  few-shot classification. In: International Conference on Learning
  Representations (2019)

\bibitem{Chen_2017_CVPR}
Chen, X., Ma, H., Wan, J., Li, B., Xia, T.: Multi-view 3d object detection
  network for autonomous driving. In: IEEE Conference on Computer Vision and
  Pattern Recognition (2017)

\bibitem{Cui_2019_CVPR}
Cui, Y., Jia, M., Lin, T.Y., Song, Y., Belongie, S.: Class-balanced loss based
  on effective number of samples. In: IEEE Conference on Computer Vision and
  Pattern Recognition (2019)

\bibitem{dai2017scannet}
Dai, A., Chang, A.X., Savva, M., Halber, M., Funkhouser, T., Nie{\ss}ner, M.:
  Scannet: Richly-annotated 3d reconstructions of indoor scenes. In: IEEE
  Conference on Computer Vision and Pattern Recognition (2017)

\bibitem{imagenet}
Deng, J., Wei, D., Richard, S., Li-Jia, L., Kai, L., Li, F.F.: Imagenet: A
  large-scale hierarchical image database. In: IEEE Conference on Computer
  Vision and Pattern Recognition (2019)

\bibitem{Dhillon_2019_arxiv}
Dhillon, G.S., Chaudhari, P., Ravichandran, A., Soatto, S.: A baseline for
  few-shot image classification. arXiv preprint arXiv:1912.03083  (2019)

\bibitem{Engelmann_2020_CVPR}
Engelmann, F., Bokeloh, M., Fathi, A., Leibe, B., Niessner, M.: 3d-mpa:
  Multi-proposal aggregation for 3d semantic instance segmentation. In: IEEE
  Conference on Computer Vision and Pattern Recognition (2020)

\bibitem{Finn_2017_ICML}
Finn, C., Abbeel, P., Levine, S.: Model-agnostic meta-learning for fast
  adaptation of deep networks. In: International Conference on Machine Learning
  (2017)

\bibitem{Clip_adapter}
Gao, P., Geng, S., Zhang, R., Ma, T., Fang, R., Zhang, Y., Li, H., Qiao, Y.:
  Clip-adapter: Better vision-language models with feature adapters. arXiv
  preprint arXiv:2110.04544  (2021)

\bibitem{He_2017_ICCV}
He, K., Gkioxari, G., Dollar, P., Girshick, R.: Mask r-cnn. In: IEEE
  International Conference on Computer Vision (2017)

\bibitem{He_2016_CVPR}
He, K., Zhang, X., Ren, S., Sun, J.: Deep residual learning for image
  recognition. In: IEEE Conference on Computer Vision and Pattern Recognition
  (2016)

\bibitem{Hou_2021_CVPR}
Hou, J., Graham, B., Niessner, M., Xie, S.: Exploring data-efficient 3d scene
  understanding with contrastive scene contexts. In: IEEE Conference on
  Computer Vision and Pattern Recognition (2021)

\bibitem{Kang_2019_ICCV}
Kang, B., Liu, Z., Wang, X., Yu, F., Feng, J., Darrell, T.: Few-shot object
  detection via feature re-weighting. In: IEEE International Conference on
  Computer Vision (2019)

\bibitem{Lang_2019_CVPR}
Lang, A.H., Vora, S., Caesar, H., Zhou, L., Yang, J., Beijbom, O.:
  Pointpillars: Fast encoders for object detection from point clouds. In: IEEE
  Conference on Computer Vision and Pattern Recognition (2019)

\bibitem{Lee_2019_CVPR}
Lee, K., Maji, S., Ravichandran, A., Soatto, S.: Meta-learning with
  differentiable convex optimization. In: IEEE Conference on Computer Vision
  and Pattern Recognition (2019)

\bibitem{Rectification}
Liu, J., Song, L., Qin, Y.: Prototype rectification for few-shot learning. In:
  European Conference on Computer Vision (2022)

\bibitem{Liu_2021_ICCV}
Liu, Z., Zhang, Z., Cao, Y., Hu, H., Tong, X.: Group-free 3d object detection
  via transformers. In: IEEE International Conference on Computer Vision (2021)

\bibitem{Lu_2021_ICCV}
Lu, Z., He, S., Zhu, X., Zhang, L., Song, Y.Z., Xiang, T.: Simpler is better:
  Few-shot semantic segmentation with classifier weight transformer. In: IEEE
  International Conference on Computer Vision (2021)

\bibitem{Ma_2021_ICCV}
Ma, J., Xie, H., Han, G., Chang, S.F., Galstyan, A., Abd-Almageed, W.:
  Partner-assisted learning for few-shot image classification. In: IEEE
  International Conference on Computer Vision (2021)

\bibitem{Min_2021_ICCV}
Min, J., Kang, D., Cho, M.: Hypercorrelation squeeze for few-shot segmentation.
  In: IEEE International Conference on Computer Vision (2021)

\bibitem{Misra_2021_ICCV}
Misra, I., Girdhar, R., Joulin, A.: An end-to-end transformer model for 3d
  object detection. In: IEEE International Conference on Computer Vision (2021)

\bibitem{Nguyen_2019_ICCV}
Nguyen, K., Todorovic, S.: Feature weighting and boosting for few-shot
  segmentation. In: IEEE International Conference on Computer Vision (2019)

\bibitem{Qi_2019_ICCV}
Qi, C.R., Litany, O., He, K., Guibas, L.J.: Deep hough voting for 3d object
  detection in point clouds. In: IEEE International Conference on Computer
  Vision (2019)

\bibitem{Qi_2017_NIPS}
Qi, C.R., Yi, L., Su, H., Guibas, L.J.: Pointnet++: Deep hierarchical feature
  learning on point sets in a metric space. In: Advances in Neural Information
  Processing Systems (2017)

\bibitem{Qiao_2021_ICCV}
Qiao, L., Zhao, Y., Li, Z., Qiu, X., Wu, J., Zhang, C.: Defrcn: Decoupled
  faster r-cnn for few-shot object detection. In: IEEE International Conference
  on Computer Vision (2021)

\bibitem{clip}
Radford, A., Kim, J.W., Hallacy, C., Ramesh, A., Goh, G., Agarwal, S., Sastry,
  G., Askell, A., Mishkin, P., Clark, J., Krueger, G., Sutskever, I.: Learning
  transferable visual models from natural language supervision. arXiv preprint
  arXiv:2103.00020  (2021)

\bibitem{Ren_2015_NIPS}
Ren, S., He, K., Girshick, R., Sun, J.: Faster r-cnn: Towards real-time object
  detection with region proposal networks. In: Advances in Neural Information
  Processing Systems (2015)

\bibitem{Sharma_2020_NIPS}
Sharma, C., Kaul, M.: Self-supervised few-shot learning on point clouds. In:
  Advances in Neural Information Processing Systems (2020)

\bibitem{Shi_2020_CVPR}
Shi, S., Guo, C., Jiang, L., Wang, Z., Shi, J., Wang, X., Li, H.: Pv-rcnn:
  Point-voxel feature set abstraction for 3d object detection. In: IEEE
  Conference on Computer Vision and Pattern Recognition (2020)

\bibitem{Shi_2019_CVPR}
Shi, S., Wang, X., Li, H.: Pointrcnn: 3d object proposal generation and
  detection from point cloud. In: IEEE Conference on Computer Vision and
  Pattern Recognition (2019)

\bibitem{Snell_17_NIPS}
Snell, J., Swersky, K., Zemel, R.S.: Prototypical networks for few-shot
  learning. In: Advances in Neural Information Processing Systems (2017)

\bibitem{Song_2022_ACL}
Song, H., Dong, L., Zhang, W.N., Liu, T., Wei, F.: Clip models are few-shot
  learners: Empirical studies on vqa and visual entailment. In: Association for
  Computational Linguistics (2022)

\bibitem{Song_2015_CVPR}
Song, S., Lichtenberg, S.P., Xiao, J.: Sun rgb-d: A rgb-d scene understanding
  benchmark suite. In: IEEE Conference on Computer Vision and Pattern
  Recognition (2015)

\bibitem{Song_2016_CVPR}
Song, S., Xiao, J.: Deep sliding shapes for amodal 3d object detection in rgb-d
  images. In: IEEE Conference on Computer Vision and Pattern Recognition (2016)

\bibitem{Sun_2021_CVPR}
Sun, B., Li, B., Cai, S., Yuan, Y., Zhang, C.: Fsce: Few-shot object detection
  via contrastive proposal encoding. In: IEEE Conference on Computer Vision and
  Pattern Recognition (2021)

\bibitem{Transformer}
Vaswani, A., Shazeer, N., Parmar, N., Uszkoreit, J., Jones, L., Gomez, A.N.,
  Kaiser, L., Polosukhin, I.: Attention is all you need. In: Advances in Neural
  Information Processing Systems (2017)

\bibitem{Vinyals_16_NIPS}
Vinyals, O., Blundell, C., Lillicrap, T., Kavukcuoglu, K., Wierstra, D.:
  Matching networks for one shot learning. In: Advances in Neural Information
  Processing Systems (2016)

\bibitem{Wang_2020_ICML}
Wang, X., Huang, T.E., Darrell, T., Gonzalez, J.E., Yu, F.: Frustratingly
  simple few-shot object detection. In: International Conference on Machine
  Learning (2020)

\bibitem{Wang_2019_ICCV}
Wang, Y.X., Ramanan, D., Hebert, M.: Meta-learning to detect rare objects. In:
  IEEE International Conference on Computer Vision (2019)

\bibitem{Wang_2020_ECCV}
Wang, Y., Fathi, A., Kundu, A., Ross, D.A., Pantofaru, C., Funkhouser, T.,
  Solomon, J.: Pillar-based object detection for autonomous driving. In:
  European Conference on Computer Vision (2020)

\bibitem{Wu_2021_ICCV}
Wu, A., Han, Y., Zhu, L., Yang, Y.: Universal-prototype enhancing for few-shot
  object detection. In: IEEE International Conference on Computer Vision (2021)

\bibitem{Wu_2021_NIPS}
Wu, A., Zhao, S., Deng, C., Liu, W.: Generalized and discriminative few-shot
  object detection via svd-dictionary enhancement. In: Advances in Neural
  Information Processing Systems (2021)

\bibitem{Wu_2020_ECCV}
Wu, J., Liu, S., Huang, D., Wang, Y.: Multi-scale positive sample refinement
  for few-shot object detection. In: European Conference on Computer Vision
  (2020)

\bibitem{Xie_2020_CVPR}
Xie, Q., Lai, Y.K., Wu, J., Wang, Z., Zhang, Y., Xu, K., Wang, J.: Mlcvnet:
  Multi-level context votenet for 3d object detection. In: IEEE Conference on
  Computer Vision and Pattern Recognition (2020)

\bibitem{Yan_2019_ICCV}
Yan, X., Chen, Z., Xu, A., Wang, X., Liang, X., Lin, L.: Meta r-cnn: Towards
  general solver for instance-level low-shot learning. In: IEEE International
  Conference on Computer Vision (2019)

\bibitem{Second}
Yan, Y., Mao, Y., Li, B.: Second: Sparsely embedded convolutional detection.
  Sensors  (2018)

\bibitem{Ye_2022_WACV}
Ye, C., Zhu, H., Liao, Y., Zhang, Y., Chen, T., Fan, J.: What makes for
  effective few-shot point cloud classification? In: IEEE Winter Conference on
  Applications of Computer Vision (2022)

\bibitem{Zhang_2021_CVPR}
Zhang, B., Xiao, J., Qin, T.: Self-guided and cross-guided learning for
  few-shot segmentation. In: IEEE Conference on Computer Vision and Pattern
  Recognition (2021)

\bibitem{Tip_Adapter}
Zhang, R., Fang, R., Zhang, W., Gao, P., Li, K., Dai, J., Qiao, Y., Li, H.:
  Tip-adapter: Training-free clip-adapter for better vision-language modeling.
  arXiv preprint arXiv:2111.03930  (2021)

\bibitem{H3DNet}
Zhang, Z., Sun, B., Yang, H., Huang, Q.: H3dnet: 3d object detection using
  hybrid geometric primitives. In: European Conference on Computer Vision
  (2020)

\bibitem{Zhao_2021_CVPR}
Zhao, N., Chua, T.S., Lee, G.H.: Few-shot 3d point cloud semantic segmentation.
  In: IEEE Conference on Computer Vision and Pattern Recognition (2021)

\bibitem{Zhou_2018_CVPR}
Zhou, Y., Tuzel, O.: Voxelnet: End-to-end learning for point cloud based 3d
  object detection. In: IEEE Conference on Computer Vision and Pattern
  Recognition (2018)

\bibitem{Zhu_2021_CVPR}
Zhu, K., Cao, Y., Zhai, W., Cheng, J., Zha, Z.J.: Self-promoted prototype
  refinement for few-shot class-incremental learning. In: IEEE Conference on
  Computer Vision and Pattern Recognition (2021)

\bibitem{cocoop}
Zhu, K., Yang, J., Li, H., Chen, C.L., Wang, X., Liu, Z.: Conditional prompt
  learning for vision-language models. In: IEEE Conference on Computer Vision
  and Pattern Recognition (2022)

\bibitem{Perceiver}
Zhu, X., Zhu, J., Li, H., Wu, X., Wang, X., Li, H., Xiaohua, W., Dai, J.:
  Uni-perceiver: Pre-training unified architecture for generic perception for
  zero-shot and few-shot tasks. In: IEEE Conference on Computer Vision and
  Pattern Recognition (2022)

\end{thebibliography}
}

%%%%%%%%%%%%%%%%%%%%%%%%%%%%%%%%%%%%%%%%%%%%%%%%%%%%%%%%%%%%
\section*{Checklist}

%%% BEGIN INSTRUCTIONS %%%
The checklist follows the references.  Please
read the checklist guidelines carefully for information on how to answer these
questions.  For each question, change the default \answerTODO{} to \answerYes{},
\answerNo{}, or \answerNA{}.  You are strongly encouraged to include a {\bf
justification to your answer}, either by referencing the appropriate section of
your paper or providing a brief inline description.  For example:
\begin{itemize}
  \item Did you include the license to the code and datasets? \answerYes{See Section~\ref{gen_inst}.}
  \item Did you include the license to the code and datasets? \answerNo{The code and the data are proprietary.}
  \item Did you include the license to the code and datasets? \answerNA{}
\end{itemize}
Please do not modify the questions and only use the provided macros for your
answers.  Note that the Checklist section does not count towards the page
limit.  In your paper, please delete this instructions block and only keep the
Checklist section heading above along with the questions/answers below.
%%% END INSTRUCTIONS %%%

\begin{enumerate}

\item For all authors...
\begin{enumerate}
  \item Do the main claims made in the abstract and introduction accurately reflect the paper's contributions and scope?
    \answerYes{}
  \item Did you describe the limitations of your work?
    \answerYes{we have included limitation discussion in the supplementary material Section A.7.}
  \item Did you discuss any potential negative societal impacts of your work?
    \answerNA{}
  \item Have you read the ethics review guidelines and ensured that your paper conforms to them?
    \answerYes{}
\end{enumerate}

\item If you are including theoretical results...
\begin{enumerate}
  \item Did you state the full set of assumptions of all theoretical results?
    \answerNA{}
        \item Did you include complete proofs of all theoretical results?
    \answerNA{}
\end{enumerate}

\item If you ran experiments...
\begin{enumerate}
  \item Did you include the code, data, and instructions needed to reproduce the main experimental results (either in the supplemental material or as a URL)?
    \answerYes{}
  \item Did you specify all the training details (e.g., data splits, hyperparameters, how they were chosen)?
    \answerYes{}
        \item Did you report error bars (e.g., with respect to the random seed after running experiments multiple times)?
    \answerYes{}
        \item Did you include the total amount of compute and the type of resources used (e.g., type of GPUs, internal cluster, or cloud provider)?
    \answerYes{}
\end{enumerate}

\item If you are using existing assets (e.g., code, data, models) or curating/releasing new assets...
\begin{enumerate}
  \item If your work uses existing assets, did you cite the creators?
    \answerYes{}
  \item Did you mention the license of the assets?
    \answerYes{}
  \item Did you include any new assets either in the supplemental material or as a URL?
    \answerYes{}
  \item Did you discuss whether and how consent was obtained from people whose data you're using/curating?
    \answerYes{}
  \item Did you discuss whether the data you are using/curating contains personally identifiable information or offensive content?
    \answerYes{}
\end{enumerate}

\item If you used crowdsourcing or conducted research with human subjects...
\begin{enumerate}
  \item Did you include the full text of instructions given to participants and screenshots, if applicable?
    \answerNA{}
  \item Did you describe any potential participant risks, with links to Institutional Review Board (IRB) approvals, if applicable?
    \answerNA{}
  \item Did you include the estimated hourly wage paid to participants and the total amount spent on participant compensation?
    \answerNA{}
\end{enumerate}

\end{enumerate}

\clearpage

\appendix

\section{Appendix}

In this supplemental material, we will include the details of dataset split in Section~\ref{sec:split}, the experimental results on base classes in Section~\ref{sec:base}, the ablation study of hard and soft assignment in Section~\ref{sec:soft}, implementation and training details of Prototypical VoteNet in Section~\ref{sec:implementation}, implementation details of the baseline method Meta VoteNet in Section~\ref{sec:implementation_meta_vote}, visualization of basic geometric primitives in Section~\ref{sec:geometric}, KNN baselines in Section~\ref{sec:knn}, non-updated prototypes in Section~\ref{sec:non}, performance on the unbalance Problem in Section~\ref{sec:performance} and limitation analysis in Section~\ref{sec:limitation}. 

\subsection{Dataset Split}

\label{sec:split}

Table \ref{tbl:data_split} lists the names of novel classes for FS-SUNRGBD and FS-ScanNet.

\begin{table}[h]
	\centering
	\scalebox{1.0}{
		\begin{tabular}{m{4.0cm} | m{4.0cm} | m{4.0cm} }\hline \toprule[0.6pt]
		~~~~~~~~~~\textbf{FS-SUNRGBD}	& ~~~~~~\textbf{FS-ScanNet}(Split-1) & ~~~~~~\textbf{FS-ScanNet}(Split-2) \\\hline
			table, bed, nightstand, toilet 
			& sofa, window, bookshelf, toilet, bathtub, garbagebin 
			& bed, table, door, counter, desk, showercurtain \\\hline
			\toprule[0.6pt]
	\end{tabular}}
	\caption{Names of novel classes for \textbf{FS-SUNRGBD} and \textbf{FS-ScanNet}.}
	\label{tbl:data_split}
\end{table} 

\subsection{Results on Base Classes}

\label{sec:base}

\begin{table*}[h]
          \centering
    \begin{tabular}{@{}l|cc|cc|cc@{}}
    \toprule
    \multirow{3}{*}{\textbf{Method}}  & \multicolumn{4}{c?}{\textbf{FS-ScanNet}} & \multicolumn{2}{c}{ \multirow{2}{*}{\textbf{FS-SUNRGBD}}}\\
    \cline{2-5}
    &  \multicolumn{2}{c?}{\textbf{Split 1}}  & \multicolumn{2}{c?}{\textbf{Split 2}} & \\
    \cline{2-7}
    & AP$_{25}$ & AP$_{50}$ & AP$_{25}$ & AP$_{50}$ & AP$_{25}$ & AP$_{50}$ \\
    \hline
    VoteNet+TFA & 51.00& 24.22 & 51.50 & 30.73 & 36.57 & 10.61\\
    VoteNet+PT+TFA & 52.78 & 25.46 & 52.04 &30.57& 38.03 & 15.44\\
    Meta VoteNet & 52.13  & 25.13& 47.89 &28.44 & 40.22& 20.60\\
    VoteNet & 57.96  & 32.60 & 54.63 & 35.76 & 47.77 & 26.78\\
    Ours & 53.83 & 28.20 & 51.22 & 32.41 & 46.07 & 25.26\\
    \hline\toprule[0.1pt]
                \end{tabular}
\caption{Results on \textbf{FS-ScanNet} and \textbf{FS-SUNRGBD} on base classes using mean Average Precision (mAP) at two different IoU thresholds of 0.25 and 0.50, denoted as AP$_{25}$ and AP$_{50}$.}
\label{tbl:base}
\end{table*}

Since we detect both base and novel classes during inference, we also report the performance on base classes in Table~\ref{tbl:base}.
For simplicity, we average the results of all k-shot (\textit{e.g.} k=1,3,5) in each split.
Table~\ref{tbl:base} shows VoteNet achieves the best performance on base classes, because it is specifically designed as a fully supervised method for base classes.
Compared with VoteNet, the performance of VoteNet+TFA and VoteNet+PT+TFA on base classes degrades by -6.96\% AP$_{25}$ and -8.38\% AP$_{50}$, and -5.18\% AP$_{25}$ and -7.14\% AP$_{50}$, respectively, on split-1 of FS-ScanNet. 
Moreover, Prototypical VoteNet retains much more ability to recognize base classes, which achieves 53.83\% AP$_{25}$ and 28.20\% AP$_{50}$ on split-1 of FS-ScanNet, and 46.07\% AP$_{25}$ and 25.26\% AP$_{50}$ on FS-SUNRGBD.
Compared with our method, we can observe a significant performance degradation in finetuning based methods. 
The reason is that, without the large-scale pre-training, a detector cannot learn more generic feature representation so as to transfer knowledge from base classes to novel classes. 
Under this circumstance, if we force the knowledge transferring by finetuning, it will distort the learned feature space for base classes.

\subsection{Hard vs. Soft Assignment}

\label{sec:soft}

\begin{table*}[h]
          \centering
    \begin{tabular}{@{}l|cc|cc@{}}
    \toprule
    \multirow{2}{*}{\textbf{Method}}  & \multicolumn{2}{c?}{\textbf{3-shot}} & \multicolumn{2}{c}{\textbf{5-shot}}\\
    \cline{2-5}
    & AP$_{25}$ & AP$_{50}$ & AP$_{25}$ & AP$_{50}$ \\
    \hline
    Soft & 27.24  & 12.82  & 29.13  & 15.13 \\
    \hline
    Hard  & 31.25 & 16.01  & 32.25  & 19.52\\
    \hline\toprule[0.1pt]
                \end{tabular}
\caption{
% Ablation study on 3-shot and 5-shot in Split-1 of \textbf{FS-ScanNet} using mean Average Precision (mAP) at two different IoU thresholds of 0.25 and 0.50, denoted as AP$_{25}$ and AP$_{50}$.
Ablation study of hard and soft assignment.}
\label{tbl:soft}
\end{table*}

We leverage the hard assignment in Section 3.2.1: Geometric Prototype Construction, in the main paper. 
Here, we compare the original hard assignment in our implemented method with the soft assignment, which calculates the similarity between a point feature with all geometric prototypes and updates all geometric prototypes in a soft manner by the similarity scores between a point feature and the geometric prototypes.
We conduct the experiment on 3-shot and 5-shot in split-1 of FS-ScanNet. 
The results in Table~\ref{tbl:soft} indicate that the hard assignment is more effective than the soft assignment. 
This is because the geometric prototypes by the hard assignment are more distinctive than those using the soft assignment, since one geometric prototype is updated using the nearest point features without considering the others.

\subsection{Implementation and Training Details}
\label{sec:implementation}

We follow recent practice~\cite{Qi_2019_ICCV,Xie_2020_CVPR} to use PointNet++~\cite{Qi_2017_NIPS} as a default backbone network. 
The backbone has 4 set abstraction layers and 2 feature propagation layers.
For each set abstraction layer, the input point cloud is sub-sampled to 2048, 1024, 512, and 256 points with the increasing receptive radius of 0.2, 0.4, 0.8, and 1.2, respectively. 
Then, two feature propagation layers successively up-sample the points to 512 and 1024.
Additionally, in both PHM and PVM, the multi-head attention layer are combined with feed forward FC layers.  
The details can be referred to our code. 
The size of memory bank of geometric prototypes is set to 120.
The number of heads in both multi-head attention networks is empirically set to 4.
The momentum coefficient is set to 0.999.
We use 40k points in FS-ScanNet and 20k points in FS-SUNRGBD as input and adopt the same data augmentation as in ~\cite{Qi_2019_ICCV}, including a random flip, a random rotation, and a random scaling of the point cloud by [0.9, 1.1].

Following the episodic training strategy~\cite{Yan_2019_ICCV}, a training mini-batch in Prototypical VoteNet is comprised of a K-shot R-class support set $D_{\rm support}$ and a R-class query set $D_{\rm train}$ (the classes in $D_{\rm support}$ and $D_{\rm train}$ is consistent). 
Each sample in $D_{\rm support}$ is the whole point cloud scene. 
Before the pooling step, we use the groud-truth bounding boxes to obtain point features of target objects. 
The network is trained from scratch by the AdamW optimizer with 36 epochs. 
The weight decay is set to 0.01. 
The initial learning rate is 0.008 in FS-ScanNet and 0.001 in FS-SUNRGBD.
Additionally, during the inference stage, we input both the query point cloud and the given support set to Prototypical VoteNet.
Note that the features of the support point cloud only need to be extracted once, as they are independent of the query feature extraction.

\subsection{Implementation Details of Meta VoteNet}

\label{sec:implementation_meta_vote}

We provide more details of the competitive baseline corresponding to Meta VoteNet in our main paper.
It is derived from an effective few-shot 2D object detection approach - Meta RCNN~\cite{Yan_2019_ICCV}. 
In Meta RCNN, each RoI feature is fused with $R$ class prototypes using the channel-wise multiplication operator.
As a result, a number of $R$ fused RoI features for each RoI are generated. 
Then each fused RoI feature is fed into the prediction layer for the binary prediction (whether the RoI feature is the category that the class prototype belongs to). 
More details can be referred to in~\cite{Yan_2019_ICCV}.
We incorporate this meta-learning approach into VoteNet, namely Meta VoteNet.
Similarly, after Sampling $\&$ Grouping (see Figure 2 in the main paper), we fuse point features of objects with $R$ class prototypes by the channel-wise multiplication operator.
Then the binary class prediction and bounding box regression are conducted based on the fused features for classification and location prediction, respectively. 
For a fair comparison, Meta VoteNet shares the same architecture with Prototypical VoteNet, except Prototypical Vote Module and Prototypical Prediction Module. 
The training details follow the proposed Prototypical VoteNet.

\subsection{Visualization of Basic Geometric Primitives}

\label{sec:geometric}

\begin{figure*}[h]
\centering
\includegraphics[width=0.8\textwidth]{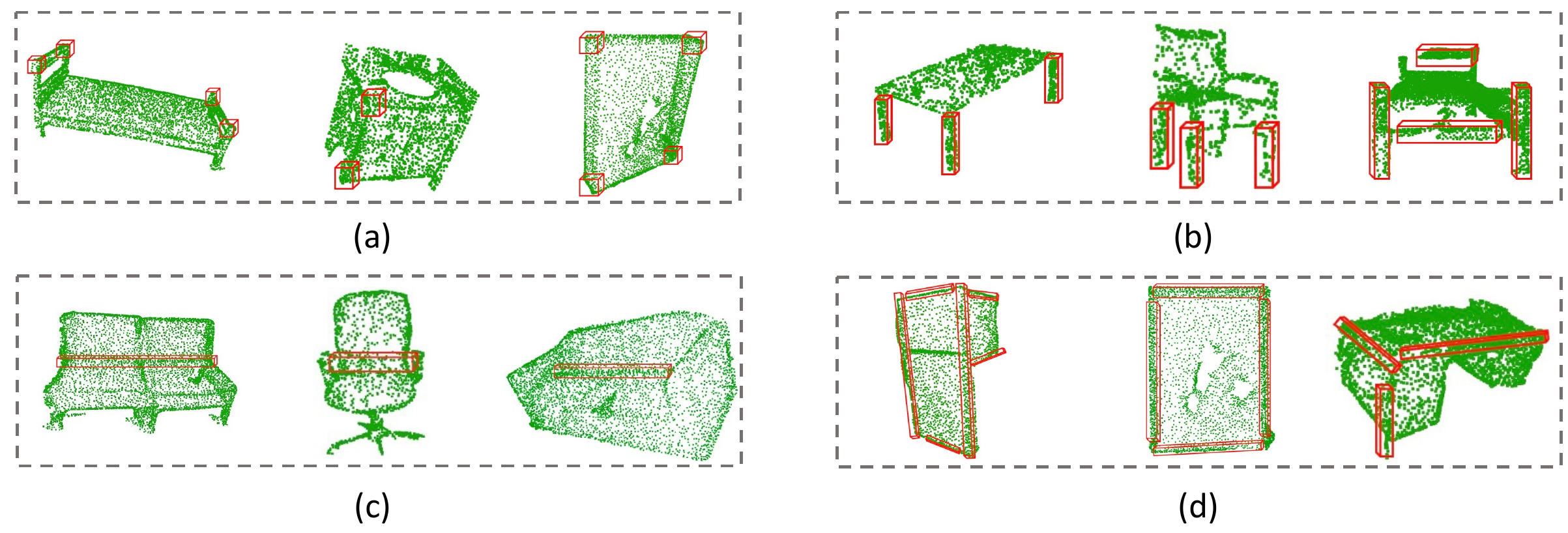}
% \vspace{-0.1in}
\caption{Visualization of some basic geometric primitives learned by the prototypes. 
(a) Corner, (b) Stick, (c) Hinge, (d) Edge. 
}
\label{fig:vis_prototype}
\end{figure*}

In Figure~\ref{fig:vis_prototype}, we visualize the relation between the learned geometric prototypes and the 3D points by searching points with features that are similar to a given geometric prototype. First, we feed object point clouds to a trained Prototypical VoteNet. Second, for each point feature, we can search for its most similar prototype. If the similarity is above a threshold, we can assign the point to that prototype. Third, we use a density-based clustering algorithm DBSCAN to cluster the point groups, and we draw the minimum 3D bounding box around each point group.
As shown in the figure, all the red bounding boxes within each subfigure belong to the same prototype. The result shows that in each subfigure, the enclosed geometric structures are similar. For example, subfigure (a) illustrates that the prototype learns the feature of corners, while subfigure (b) shows that the prototype learns the long stick.

\subsection{KNN Baselines}

\label{sec:knn}

\begin{table*}[h]
	\centering
	{
		\begin{tabular}{@{}l|cc|cc@{}}
    \toprule
    \multirow{2}{*}{\textbf{Method}}  & \multicolumn{2}{c?}{\textbf{3-shot}} & \multicolumn{2}{c}{\textbf{5-shot}}\\
    \cline{2-5}
    & AP$_{25}$ & AP$_{50}$ & AP$_{25}$ & AP$_{50}$ \\
    \hline
    VoteNet + KNN & 23.07 & 9.56 & 25.58 & 13.51 \\
\hline
    Group-Free + KNN & 24.22 & 9.97 & 26.33 & 13.92 \\
\hline
    3DETR[2] + KNN  & 24.08 & 10.21 & 26.01 & 14.36 \\
\hline
    Ours &  31.25  & 16.01  & 32.25  & 19.52\\
    
    \hline\toprule[0.1pt]
\end{tabular}}
    \caption{KNN baselines.}
	\label{tbl:knn}
\end{table*}

We apply KNN assignment to VoteNet and two SOTA 3D detectors GroupFree~\cite{Liu_2021_ICCV} and 3DETR~\cite{Misra_2021_ICCV}. We conducted this experiment on 3-shot and 5-shot in split-1 of FS-ScanNet. The KNN assignment is realized by calculating the distance between each object feature and features of all training objects in the classification step, and assigning the sample to the class based on voting from its k-nearest objects of the training set. Here, we take k as one since we find increasing the value k doesn’t improve performance. The results are shown in Table~\ref{tbl:knn}.
Comparing the performance of “VoteNet + KNN” and “ours”, we see that the non-parametric KNN classifier will not help improve few-shot learning much (“VoteNet” vs “VoteNet+KNN”).
Comparing the performance of different detectors (“VoetNet+KNN”, “GroupFree + KNN”, and “3DETR+KNN”), we observe that a better detection architecture does not bring large performance gains in the few-shot 3D detection scenario.
The most challenging issue for few-shot 3D object detection still lies in how to learn effective representation if only a few training samples are provided. The classifier and architecture don’t help much if the model cannot effectively extract features to represent novel categories with only a few samples.

\subsection{Non-Updated Prototypes}

\label{sec:non}

\begin{table*}[h]
	\centering
	{
		\begin{tabular}{@{}l|cc|cc@{}}
    \toprule
    \multirow{2}{*}{\textbf{Method}}  & \multicolumn{2}{c?}{\textbf{3-shot}} & \multicolumn{2}{c}{\textbf{5-shot}}\\
    \cline{2-5}
    & AP$_{25}$ & AP$_{50}$ & AP$_{25}$ & AP$_{50}$ \\
    \hline
    Update & 28.05  & 13.89  & 28.51  & 14.51 \\
\hline
    Non-Update & 31.25 & 16.01 & 32.25 & 19.52 \\
    
    \hline\toprule[0.1pt]
\end{tabular}}
    \caption{Does setting the prototype at the end (no updates) perform well ?}
	\label{tbl:non-update}
\end{table*}

As shown in Table~\ref{tbl:non-update}, for the proposed Prototypical VoteNet, if we don’t update the prototype in PVM, the performance would degrade significantly. Without updating, the randomly initialized prototypes can not learn the geometry information from base classes in the training phase. In this case, it is hard to transfer the basic geometry information from base classes to the novel classes as the prototypes are meaningless.

\subsection{Performance on the Unbalance Problem}

\label{sec:performance}

\begin{table*}[h]
	\centering
	{
		\begin{tabular}{@{}l|cc|cc|cc|cc}
    \toprule
    \multirow{2}{*}{\textbf{Method}}  & \multicolumn{2}{c?}{\textbf{P}} & \multicolumn{2}{c?}{\textbf{10P}} & \multicolumn{2}{c?}{\textbf{25P}} & \multicolumn{2}{c}{\textbf{50P}}\\
    \cline{2-9}
    & AP$_{25}$ & AP$_{50}$ & AP$_{25}$ & AP$_{50}$ & AP$_{25}$ & AP$_{50}$ & AP$_{25}$ & AP$_{50}$ \\
    \hline
    VoteNet & 62.34 & 40.82 & 52.06	& 35.64	& 43.12	& 27.13 &	40.01 &	26.77 \\
\hline
    Ours & 62.59 &	41.25 &	52.60 &	36.87 &	44.53 &	29.17 &	41.99 &	29.01 \\
    \hline\toprule[0.1pt]
\end{tabular}}
    \caption{Performance on the Unbalance Problem in ScanNet V2}
	\label{tbl:un-scan}
\end{table*}

\begin{table*}[h]
	\centering
	{
		\begin{tabular}{@{}l|cc|cc|cc|cc}
    \toprule
    \multirow{2}{*}{\textbf{Method}}  & \multicolumn{2}{c?}{\textbf{P}} & \multicolumn{2}{c?}{\textbf{10P}} & \multicolumn{2}{c?}{\textbf{25P}} & \multicolumn{2}{c}{\textbf{50P}}\\
    \cline{2-9}
    & AP$_{25}$ & AP$_{50}$ & AP$_{25}$ & AP$_{50}$ & AP$_{25}$ & AP$_{50}$ & AP$_{25}$ & AP$_{50}$ \\
    \hline
    VoteNet & 59.78	& 35.77	& 51.09 &	31.81 &	43.68 &	29.08 &	40.46 &	22.23 \\
\hline
    Ours & 60.34 &	36.80 &	51.85 &	32.98 &	44.66 &	31.93 &	41.84 &	25.04 \\
    \hline\toprule[0.1pt]
\end{tabular}}
    \caption{Performance on the Unbalance Problem in SUN RGB-D}
	\label{tbl:un-sun}
\end{table*}

To analyze the performance of the proposed model on the imbalance problem, we conduct experiments using all the classes. Note that we conduct the experiments not only on ScanNet V2, but also on the more unbalanced counterparts. 
We follow the benchmark~\cite{Cui_2019_CVPR}, to create these counterparts: 1) sorting the classes in descending order according to number of samples in each class, then we have $n_i > n_j$ if $i < j$, where $n$ is the number of samples, $i$ and $j$ denote the index of the classes. 2) reducing the number of training samples per class according to an exponential function $n=n_i*u^i$, where $u \in (0,1)$. The test set remains unchanged. 
According to the benchmark~\cite{Cui_2019_CVPR}, we define the imbalance factor of a dataset as the number of training samples in the largest class divided by the smallest. Note that we use P as the value of the imbalance factor in ScanNet V2. Additionally, we add another three sets, whose values of imbalance factor are 10P, 25P and 50P for both ScanNet V2. 
As shown in Table~\ref{tbl:un-scan}, we achieve comparable performance in the original dataset setting. With the imbalance becoming more severe (e.g., 25P, 50P), our approach outperforms the baseline more.
Note that our focus is on few-shot 3D object detection, where representation learning of new categories becomes the top consideration of algorithm design. This few-shot problem is more useful for scenarios where many new categories appear frequently and require the system to quickly adapt to recognize them.
However, the long-tailed problem focuses on how to learn good representations and classifiers that can deliver good performance for both head and tail categories. We believe that dedicated designs can further improve the performance of long-tailed 3D object detection. We will also add the results and analysis for the long-tailed setting in our paper and hope to inspire more future investigations.

\subsection{Limitation Analysis}

\label{sec:limitation}

Although the 3D cues of point clouds are more stable since they can get rid of some visual distractors, such as lighting and perspectives, some factors still impede the model from better generalization. For instance, in 3D scene understanding, if the point cloud in the training set is dense and that of the test set is sparse, a model often performs poorly, which can be treated as a cross-domain problem. Regarding few-shot 3D object detection, the performance might degrade if there is such a large domain gap between base classes and novel classes. Even though the basic geometric features are learned in the base classes, they might not be generalized well to the novel classes due to the difference in point cloud sparsity. The performance of this model has much room for improvement. One way to achieve better performance is large-scale pre-training. Large-scale pre-training enables the model to learn more generic features for transfer learning using limited samples, which benefits the community of 2D few-shot learning (i.e., ImageNet Pre-training). For future works, we might resort to the pre-training models in the 2D domain to facilitate the few-shot generalization on 3D few-shot learning and how these techniques can be combined with our method.

\end{document}